\documentclass{article}

\usepackage{microtype}
\usepackage{graphicx}
\usepackage{booktabs} 

\usepackage{hyperref}

\usepackage[accepted]{icml2021}

\usepackage[utf8]{inputenc} 
\usepackage[T1]{fontenc}    
\usepackage{url}            
\usepackage{booktabs}       
\usepackage{amsfonts}       
\usepackage{nicefrac}       
\usepackage{microtype}      
\usepackage{color}
\usepackage{graphicx}
\usepackage{bbm}
\usepackage{dsfont}
\usepackage{amsmath}
\usepackage{subcaption}
\usepackage{multirow}
\usepackage{lipsum}
\usepackage{xcolor}
\usepackage{sidecap}
\newcommand{\bla}{\color{black}}
\newcommand{\blu}{\color{black}}
\newcommand{\sun}[1]{{\color{black} #1}}

\DeclareMathOperator*{\argmax}{arg\,max}
\DeclareMathOperator*{\argmin}{arg\,min}
\newtheorem{theorem}{Theorem}
\newtheorem{assumption}{Assumption}

\icmltitlerunning{Model Distillation for Revenue Optimization: Interpretable Personalized Pricing}

\begin{document}

\twocolumn[
\icmltitle{Model Distillation for Revenue Optimization: \\Interpretable Personalized Pricing}



\icmlsetsymbol{equal}{*}

\begin{icmlauthorlist}
\icmlauthor{Max Biggs}{vr}
\icmlauthor{Wei Sun}{ibm}
\icmlauthor{Markus Ettl}{ibm}
\end{icmlauthorlist}

\icmlaffiliation{vr}{Darden School of Business, University of Virginia, Virginia, USA.}
\icmlaffiliation{ibm}{IBM Research, Yorktown Heights, New York, USA}

\icmlcorrespondingauthor{Max Biggs}{biggsm@darden.virginia.edu}

\icmlkeywords{Prescriptive analytics, decision making, interpretability, pricing}

\vskip 0.3in
]



\printAffiliationsAndNotice{}  

\begin{abstract}
 Data-driven pricing strategies are becoming increasingly common, where customers are offered a personalized price based on features that are predictive of their valuation of a product. It is desirable for this pricing policy to be simple and interpretable, so it can be verified, checked for fairness, and easily implemented. However,  efforts to incorporate machine learning into a pricing framework often lead to complex pricing policies which are not interpretable,  resulting in slow adoption
 in practice. We present a customized, prescriptive tree-based algorithm that distills knowledge from a complex black-box machine learning algorithm, segments customers with similar valuations and prescribes prices in such a way that maximizes revenue while maintaining interpretability. We quantify the regret of a resulting  policy and demonstrate its efficacy in applications with both synthetic and real-world datasets.\end{abstract}

\section{Introduction}

 To remain competitive, many firms are seeking to implement data-driven pricing policies. This has become possible due to the large amounts of relevant data firms are collecting, and through recent advances in 
machine learning. 
This data rich environment has also enhanced personalized pricing, where customers are offered a personalized price based on features that are predictive of their valuation of a product. 
A common example of personalized pricing is targeted offers through loyalty programs. 
While personalized pricing may potentially lead to higher revenue, currently only a quarter of retail companies are supportive of ML adoption, the lowest acceptance rate across surveyed industries.\footnote{https://advisory.kpmg.us/content/dam/advisory/en/pdfs/2020/ retail-living-in-an-ai-world.pdf} One of the main concerns is that the most accurate prediction models  are often non-parametric functions which are considered as a black-box.  Their opaqueness makes it difficult for firms 
to understand and trust the underlying mechanics of how the prices are being formed, which may potentially lead to fairness and legal issues. Another challenge with a personalized pricing policy is that it might be very  complex, making it difficult and cumbersome for a firm to implement and to verify the prices are sensible.

\sun{In the recent years, there has been a surge of interest in making machine learning models more interpretable (\citealp{letham2015interpretable, ustun2016supersparse, lakkaraju2016interpretable, bertsimas2019price}). A popular approach to gain interpretability is  knowledge distillation   \cite{hinton2015distilling}, where a simple model (the \emph{student}) is trained on a transformed dataset, representing the distilled knowledge extracted by a complex model (the \emph{teacher}). The distillation process often shows to bring the student closer to matching the predictive power of the teacher. Among many potential candidate models as the student (e.g., linear models and generalized additive models), decision tree stands out due to its ability to capture interactions between features, scalability to large datasets, and perhaps most importantly, providing ``human-friendly'' explanations to aid interpretability \cite{frosst2017distilling,bastani2017interpreting,bastani2018interpreting,miller2019explanation}.    }

\sun{However, the bulk of existing work on interpretability  focuses on  predictive models. There has been significantly less work on interpreting policies from these models when embedded in an operational decision-making context, which is the topic of this paper. It is crucial to note that it is not straightforward to adapt predictive models effectively to a prescriptive setting. 
Consider the problem of extracting interpretable pricing rules under a revenue-maximizing objective; a potential method of adapting the student-teacher framework is to first train a black-box teacher model that predicts demand (i.e., a response function that relates sales to price and other relevant features) and use this model to determine the best price with respect to the given features. Next, one  can then  approximate these prices by training an interpretable student model, e.g., regress on the prices with a decision tree. In the appendix \footnote{https://arxiv.org/abs/2007.01903}, we provide a simple  example which illustrates the pitfalls of this naive application of the student-teacher framework in the prescriptive setting. The key insight is that a partition of data which leads to good predictive accuracy (e.g.,  approximating given prices) does not necessarily translate to good prescriptive decisions (e.g., maximize revenue). 
Pricing data also has the challenge that past pricing decisions are made based on customer features and their perceived willingness to pay. Making decisions without properly controlling for these effects leads to sub-optimal revenue, as  documented in \citealp {bertsimas2016power}.}

\begin{figure*}
  \centering
  \includegraphics[width=0.8\linewidth]{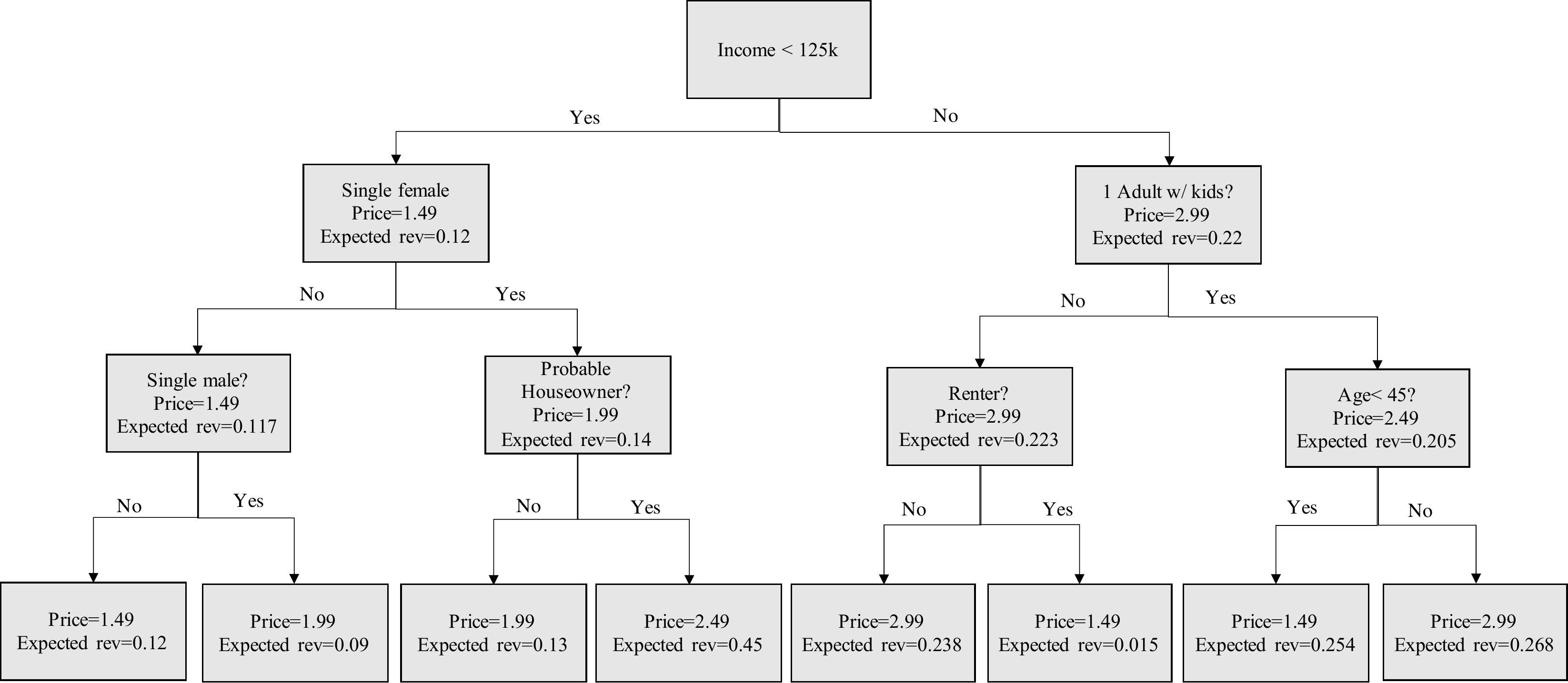}
  \caption{Interpretable personalized pricing algorithm for strawberries from Dunnhumby dataset}
  \label{pricing_policy_st}
\end{figure*}


To address these issues, we propose an interpretable method for \blu generating personalized pricing policies\bla, by adapting the student-teacher framework specifically for a decision-making setting and introducing a new tree algorithm – prescriptive student trees. To achieve this, we first train an accurate but opaque predictive teacher model, such as Boosted Trees or Neural Networks \blu to predict the demand\bla. Using this model we are able to produce an optimal price for each item, but the pricing policy is often complex and not interpretable. \sun{Instead of regressing on prices as in the naive approach discussed earlier, }the prescriptive student model seeks to distill this pricing policy to one which is simpler and more interpretable through a customized recursive partitioning algorithm.  Each leaf in the prescriptive student tree defines a set of items which are given the same price. These items share similarities in their covariates, and are chosen in such a way that each item in a leaf has a similar optimal price, as evaluated by the teacher model. In other words, the role of the teacher in the prescriptive setting is to estimate the counterfactual outcomes that are missing from the data, and control for observed confounding variables in the student tree construction process. 

One example of an interpretable pricing policy produced by our algorithm is shown in Figure \ref{pricing_policy_st}, with eight segments. This particular example is a personalized pricing policy for strawberries and is explored further in the grocery pricing case study in Section~\ref{dunnhumby_case_study}. Each segment is assigned a price and the expected revenue per shopper is also shown. As an example, it suggests \$1.99 for a single male with income below \$125k, but \$2.99 for an older shopper with kids and an income above \$125k.

Our key contributions are as follows: we propose a method for prescribing interpretable policies, where  we introduce a new impurity measure and algorithm for constructing a student prescriptive tree by incorporating a teacher model. We quantify the regret of the resulting  policy and demonstrate its efficacy in applications with both synthetic and real-world datasets. We want to point out that while we focus on revenue optimization as the running example, this method can be easily extended to many other prescriptive settings, such as precision medicine and personalized recommendation. More details will follow in Section \ref{sect_method}.  

\section{Related literature}

\sun{Popularized by \citealp{hinton2015distilling}, in the vanilla version of the knowledge distillation approach, the student model is trained on a dataset where the observed outcomes are replaced by soft-labels predicted by the teacher model. }
Variations on this idea have been extensively studied in deep learning and for other non-parametric models, for example, SHAP: \citealp{lundberg2017unified}, LIME: \citealp{ribeiro2016should}, Anchors: \citealp{ribeiro2018anchors}, RuleFit: \citealp{friedman2008predictive}, also \citealp{baehrens2010explain, sanchez2015towards, bucilua2006model}. Of particular interest to this work is \citealp{craven1996extracting, bastani2018interpreting}, which use decision trees as the approximating model. Rather than using a teacher-student framework to approximate a prediction model, we adapt it to approximate a prescriptive policy.

There has been a recent focus within the operations research community on prescriptive analytics, i.e., how to use data and machine learning to make better operational decisions (\citealp{ban2019big, bertsimas2020predictive, biggs2017optimizing, mivsic2017optimization, bertsimas2018optimization, anderson2020strong}). This includes personalized pricing models, where user and/or product data is used to customize prices for products (\citealp{chen2015statistical, ferreira2016analytics, javanmard2016dynamic, bertsimas2016power, ye2018customized, baardman2018detecting, elmachtoub2018value, ban2020personalized}). However, these approaches are not explicitly concerned with interpretability, and indeed may result in complex decision policies with no segmentation or limit on the number of prices offered.

We highlight some tree-based prescriptive approaches which are closest to the current work. \citealp{kallus2017recursive} introduce personalization trees, a customized tree algorithm which proposes an impurity measure based on prescribing the treatment which has the highest outcome on average within each set in the partition. \citealp{bertsimas2019optimal} extend this approach by incorporating predictive accuracy into the tree construction algorithm, and exploring more sophisticated optimization techniques. \citealp{athey2016recursive} introduce a causal tree algorithm for estimating heterogenous treatment effects for binary treatments.  \citealp{zhou2018offline} apply a doubly robust estimation approach to estimating reward of a policy. \citealp{ciocan2018interpretable} show a tree-based approach for solving optimal stopping problems. \citealp{elmachtoub2020decision} recently provides a tree-based implementation of the predict-then-optimize approach first introduced in \citealp{elmachtoub2017smart}. 
While this is a promising approach, it is not clear how to apply this to the pricing setting where the parameter being estimated is not just a function of exogenous covariates, but also price which is the decision being made.  

One implicit assumption behind the existing prescriptive tree-based approaches \citep{athey2016recursive, kallus2017recursive, bertsimas2019optimal} is that the leaves are essentially homogeneous, so the \blu confounding \bla effects of the other \blu (observed) \bla variables are controlled for. This assumption means that the only difference between outcomes in the same leaf is due to the treatment assigned. This might be a reasonable assumption when the trees are very deep (i.e., resulting in smaller leaves where the covariates are similar, risking overfitting), but is unlikely to hold in shallow trees. \blu In many prescriptive settings, shallow trees with fewer segments which translate into simpler pricing policies are desirable. \bla 
By incorporating a teacher model to guide the pricing policy, our method controls for the observed confounding variables at any depth, rather than assuming they are the same, therefore ensuring that confounding effects are minimized. We also note that these approaches only work for discrete treatments, \blu whereas our approach can be applied to continuous treatments. \bla

 
\section{Methodology}\label{sect_method}
\subsection{Problem formulation
}

We assume we have $n$ observational data samples $\{(x_i,p_i,y_i)\}_{i=1}^n$, where $x_i \in \mathcal{X}^d$ are features which describe the $i^{th}$ item, $p_i \in R$ is the price assigned to the item, and $y_i \in \{0,1\}$ is whether the item sold $(1)$ or not $(0)$. \blu For convenience of exposition, we refer to ``item'' as a selling unit of a product, where $x_i$ may only contain product features as in the airline pricing case study in Section~\ref{sect:airline}. It may also represent a  (product, customer) pair when $x_i$ also includes  customer features, as in the Dunnhumby grocery store case study in Section~\ref{dunnhumby_case_study}.
\bla

\blu As is common practice in the causal inference literature, we make the ignorability assumption \cite{hirano2004propensity}, that is, there were no unmeasured confounding variables that affected both the choice of price and the outcome. \bla Accordingly, we assume an underlying function $f:\mathcal{X}^d \times R \rightarrow [0,1]$ which maps the item features and price to a purchase probability $f(x,p)=E[Y|X=x,P=p]=\mathds{P}[Y=1|X=x,P=p]$. An optimal \textit{personalized pricing} algorithm selects a price $\tau^*(x)$  for each item based on its features  to maximize the item's expected revenue:  
\begin{equation}
\label{true_rev}
\tau^*(x)= \argmax_{p \in R} p f(x,p) 
\end{equation}

In reality, the true response function $f(x,p)$ is unknown, but can be estimated by solving an empirical risk minimization problem $\hat{f}=\argmin_{f \in \mathcal{F}} \sum_{i}^n L(x_i,p_i,y_i; f)$ over a class of functions $\mathcal{F}$ and an appropriate loss function $L$ for classification. \blu  
We refer to $\hat{f}$ as the teacher model. 
\bla

If we substitute the proxy response function $\hat{f}$ into the optimization in (\ref{true_rev}), the resulting personalized pricing policy may be too complex for a firm to verify and implement, with each feature vector potentially resulting in a different price. Furthermore, if the class of functions $\mathcal{F}$ is complex, there may not be any insight into how the predictions are made. Our goal is to find an \textit{interpretable personalized pricing} policy. We achieve this by restricting the pricing policy to the class of decision tree functions of depth $ k, ~ \mathcal{T}(k)$, \blu which we refer to as the prescriptive student model, \bla and find the policy which achieves the highest revenue as predicted by the teacher model:
\begin{equation}
\label{empirical_price_opt}
\max_{\tau(x) \in \mathcal{T}(k)}   \frac{1}{n}\sum_{i=1}^n \tau(x_i) \hat{f}(x_i,\tau(x_i))
\end{equation}
A decision tree (see \citealp{breiman1984classification}) is a class of functions which recursively partitions the feature space into leaves, each of which is associated with a prediction. The binary tree structure is defined by internal nodes and leaves. At each internal node there is an axis-aligned split which checks if a feature is less than a threshold. If this condition is met when mapping a data point to a leaf then it proceeds to the left child node. Otherwise, it continues to the right child node until a leaf is reached.

Contrary to the well-studied predictive setting where each leaf has a prediction of the outcome, the leaves of the prescriptive tree specify a single price at which to sell the items. Decision trees result in interpretable pricing policies because specifying the depth of the tree restricts the number of unique prices in the policy. Furthermore, the set of items which are prescribed a price is defined by the single variable splits which lead to the leaf.  Each of these rules can be easily verified by a practitioner. An example of a tree-based pricing policy is shown in Figure \ref{pricing_policy_st}.

While we focus on pricing, the framework we propose can be easily adapted to broad application in prescriptive settings with a very minor change to Eq \ref{true_rev} from maximizing revenue $ p f(x,p)$ to maximizing any known function of a predictive model $ g(f(x,p))$. Consider a Warfarin dosing application in personalized medicine \cite{kallus2017recursive,bastani2020mostly}. We can define a criterion to maximize the probability that a correct dosage is given; for personalized recommendations for a job training program \cite{kallus2017recursive}, the criterion can be to maximize the net income after enrollment cost.

\subsection{Approximation guarantees} 
We show that the regret of a tree-based interpretable pricing policy optimized over the teacher model can be bounded relative to the optimal personalized pricing policy. Here the regret is defined as the difference in the expected revenue between the two policies.  To do so, we assume uniform bounds on the error of the estimated teacher model and that the true expected revenue function is smooth. For convenience of exposition, we also assume $x \in [0,1]^d$.  

 \begin{assumption}
 \label{uniform_assumption}
Estimation error uniformly bounded:$|f(x,p) - \hat{f}(x,p)| \leq \mathcal{K}(n), \forall x \in [0,1]^d, p \in R $
\end{assumption}

An example where assumption \ref{uniform_assumption} is met is Honest trees and forests from \citealp{wager2018estimation}.

\begin{assumption}
$pf(x,p)$ is Lipshitz continuous in $x$ with respect to the $L_2$ norm: $$pf(x,p)-pf(x',p) \leq L \|x- x' \|_2, ~~ \forall x,x' \in [0,1]^d$$
\end{assumption}


\begin{theorem}
Under the above assumptions, there exists a tree-based policy $\hat{\tau}(x) \in \mathcal{T}(k)$, trained on the proxy 
response function $\hat{f}$ such that:
\begin{align}
\tau^*(x)f(x,\tau^*(x)) - &\hat{\tau}(x)f(x,\hat{\tau}(x)) \leq \\ 
& 2^{\frac{-k}{d}+1} L \sqrt{d}   +  2 \mathcal{K}(n)   ~~ \forall  x \in [0,1]^d \nonumber
\end{align}	
\end{theorem}

The proof is given in the appendix.  This theorem shows that the revenue gap between the tree-based pricing policy and the optimal personalized policy can be decomposed into an approximation error and an estimation error. The estimation error is inherited from the estimation error of the teacher model. The approximation error depends on the smoothness of the function and the dimension of the feature space, and decays rapidly with the depth of the tree.  


\subsection{Recursive partitioning algorithm for student prescriptive trees}

We propose a simple algorithm for solving (\ref{empirical_price_opt}).
Considering the task of constructing an optimal classification tree is known to be NP-complete \cite{laurent1976constructing}, we use a heuristic based on the recursive partitioning \cite{breiman1984classification}, whereby splits are chosen greedily. We achieve this by proposing a new splitting criterion, which utilizes the teacher model. Denote $S_l \subseteq [n]$ as the subset of observations which belong to leaf $l$ of the student tree. We define the \textit{revenue maximization criterion} as follows:
\begin{equation}
\label{eq:rev_max_student}
	\mathcal{R}(S_l)= \max_p \sum_{i \in S_l} p \hat{f}(x_i,p)
\end{equation}
At each node in the tree, we consider a decision split partitioning the data into two sets $S_1(j,s)= \{i \in [n] | x_{i,j} \leq  s \}$ and $S_2(j,s)= \{i \in [n] | x_{i,j} >  s \}$, where $x_{i,j}$ is the $j^{th}$ feature of observation $i$. We choose the split which results in maximum predicted revenue by giving each set formed a different price. We do this by optimizing over $j$, the feature chosen to split on, and $s$, the threshold, i.e., $\max_{j,s} \mathcal{R}(S_1(j,s)) + \mathcal{R}(S_2(j,s))$.

 On the training data this is limited to searching the $d$ dimensions of the feature vector, while there are at most $n$ unique splits in the data, one for each data point. We continue recursively splitting until a specified criteria has been met. We use tree depth as termination criteria which corresponds to the desired complexity of the pricing policy,  but can easily incorporate other commonly used approaches such as minimum leaf size criterion, or minimum improvement in impurity/revenue. In each leaf, the price assigned is the maximizer of (\ref{eq:rev_max_student}).

\blu In many pricing applications, there is a discrete set of prices that can be chosen from,  $ p \in \bar{\mathcal{P}}=\{p_1,p_2,...,p_m\}$.  \bla
With this simplification, the estimated revenue for each item at each price $r_{i,k}=p_k \hat{f}(x_i,p_k), k \in [m]$ can be calculated prior to tree construction. The revenue for each segment can be calculated and used instead of  (\ref{eq:rev_max_student}):
\begin{equation}
\label{eq:splitting_proceedure}
	\mathcal{R}_m(S_l)= \max_{k \in [m] } \sum_{i \in S_l} r_{i,k}
\end{equation}
We note that there are non-greedy approaches for constructing trees which could be adapted to solve (\ref{empirical_price_opt}), such as \citealp{bertsimas2019optimal}. However, experiments in \citealp{bastani2018interpreting} have shown that greedy trees are more intuitive as the variables leading to the largest improvement are split on first, matching human reasoning and providing more interpretabilty. Furthermore greedy algorithms are known to scale better and solve more quickly for large datasets.

\begin{table*}[]
\centering
\caption{Revenue comparison on synthetic datasets}
\begin{tabular}{ll lll lll lll}
\hline
 & & \multicolumn{3}{c}{Mean revenue} &\multicolumn{3}{c}{Maximum revenue}&\multicolumn{3}{c}{Minimum revenue} \\
 \cmidrule(lr){2-5} \cmidrule(lr){6-8} \cmidrule(lr){9-11}
Data &  Optimal &  PT   & SPT   & CT    &  PT   & SPT   & CT    &  PT   & SPT   & CT    \\
\hline
1 & 3.28 & 3.07 & \textbf{3.19} & 3.13 & 3.24          & \textbf{3.27} & 3.26 & 2.88 & \textbf{3.01} & 2.99 \\
2 & 2.94 &	1.75&	\textbf{2.14}&	1.70&	1.93 &	\textbf{2.27}&	1.79&	1.58&	\textbf{1.95}&	1.62\\
3 & 3.42&	3.20&	\textbf{3.34}&	3.21&	3.36&	\textbf{3.42}&	3.37&	2.78&	\textbf{3.24}&	3.02 \\
4 &3.49&	3.13&	\textbf{3.36}&	3.24&	3.29&	\textbf{3.46}&	3.33&	2.82&	\textbf{3.20}&	3.08 \\
5 & 3.35&	3.11	&\textbf{3.26}&	3.06&	3.29&	\textbf{3.33}	&3.26&	2.79&	\textbf{3.12}&	2.47\\
6& 2.58	&2.01&	\textbf{2.40}&	2.12&	2.34&	\textbf{2.48}&	2.41&	1.53	&\textbf{2.22}&	1.87\\
\hline
\end{tabular}
\label{revenue_table}
\end{table*}

\section{Experimental results}

\subsection{Benchmarks}\label{existing_approaches}
We compare our approach with two other state-of-the-art prescriptive tree-based algorithms.  
Personalization trees \cite{kallus2017recursive}, prescribe the treatment which has the highest average outcome in each leaf. This algorithm can be adapted to a personalized pricing setting by using the observed revenue as the outcome for each item.  
More specifically, if there are $m$ prices $t \in [m]$, we calculate the impurity for a set of items $S_l$ as follows, and use this as the splitting criterion in a greedy algorithm: 
\begin{equation}
\label{kallus_impurity}
\mathcal{I}(S_l) =  \max_{t \in [m]} \frac{\sum_{i \in S_l} p_i y_i \mathds{1}\{p_i=t\} }{\sum_{i \in S_l} \mathds{1}\{p_i=t\}  } 
\end{equation}
We also benchmark our algorithm against the causal trees algorithm from \citealp{athey2016recursive}. The  approach estimates heterogeneous treatment effects, rather than prescribing the optimal treatment. In particular, it only considers estimation of CATE (conditional average treatment effect) for binary treatments $\delta(x)=E[Y|X=x,T=1]-E[Y|X=x,T=0]$. Following the testing procedure in \citealp{kallus2017recursive}, we adapt the causal tree algorithm to prescribing the optimal treatment from  a discrete set of alternatives using the  ``one-vs.-all" approach. In this approach, the treatment effect of each price is estimated by comparing the expected outcome of those items which received the price to those which received another price. Specifically, for each price $t \in [m]$ , the treatment effect $ \delta(x,t)^{1vall}= E[Y|X=x,P=t]-E[Y|X=x,P \neq t]$ is estimated using a causal tree and the prescribed treatment is the one which achieves the highest improvement, i.e., $\max_{t \in [m]} \delta(x,t)^{1 v all} $. Note that this approach is no longer as interpretable as a single tree, given that there is no guarantee that the tree partitions associated with each treatment will align, resulting in a loss of the recursive tree structure and significantly more complex policies 

To train the causal trees, we use the implementation from \citealp{athey2016recursive}, using default parameter values. Due to the lack of open-source code for \citealp{kallus2017recursive}, we implement the impurity measure (\ref{kallus_impurity}) in our own recursive partitioning procedure. 

 
 


\subsection{Results on simulated datasets}

Simulated datasets allow us to accurately evaluate the counterfactual outcomes associated with changing prices, and calculate the resulting revenue from a policy, since the underlying probability model of simulated dataset is known. 
\begin{figure}[hbt!]
\begin{subfigure}{.47\textwidth}
  \includegraphics[width=1\textwidth]{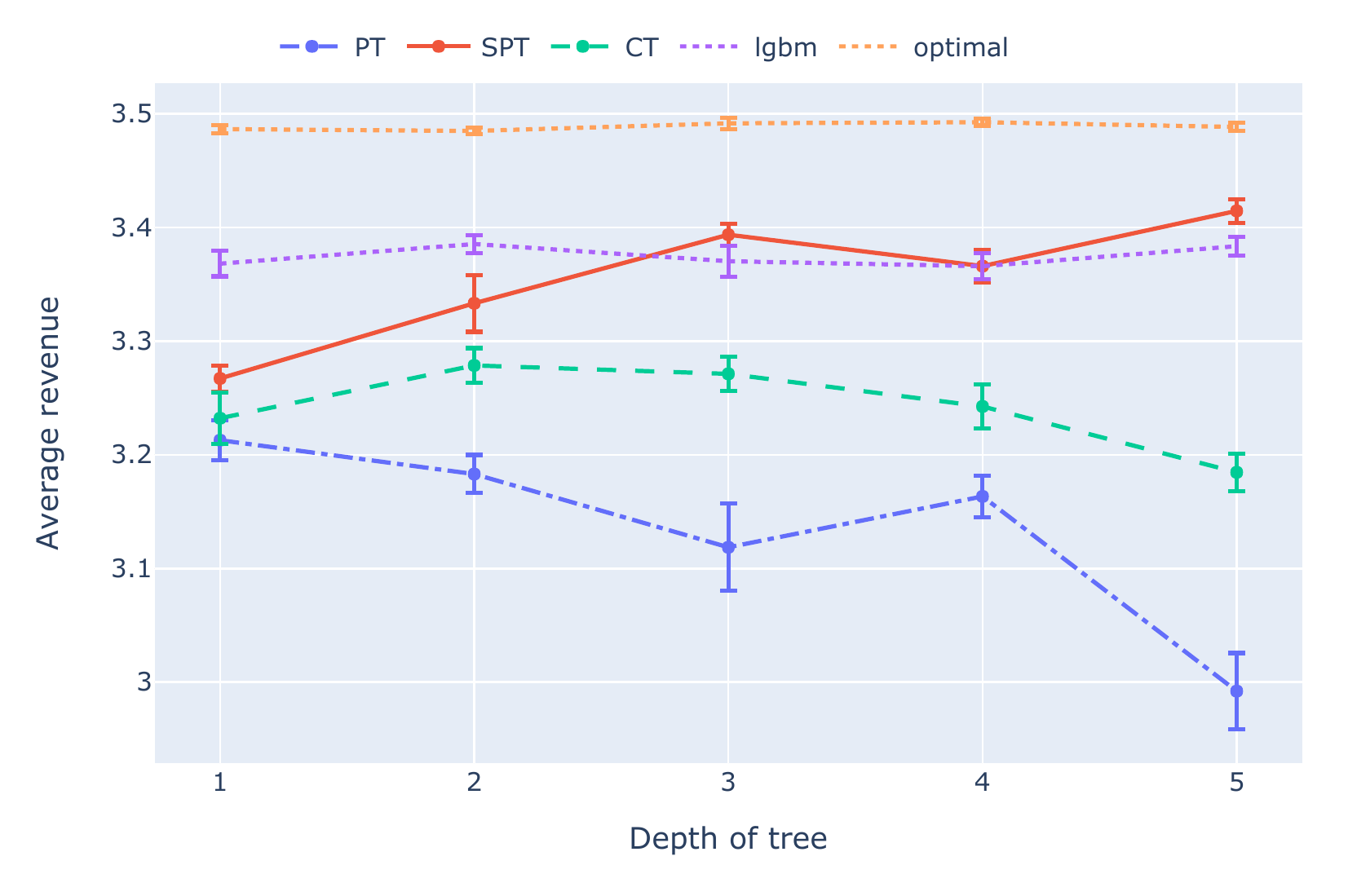}
  \caption{Revenue with varying tree depth}
  \label{rev_with_depth}
  \end{subfigure}
\begin{subfigure}{.47\textwidth}
  \includegraphics[width=1\textwidth]{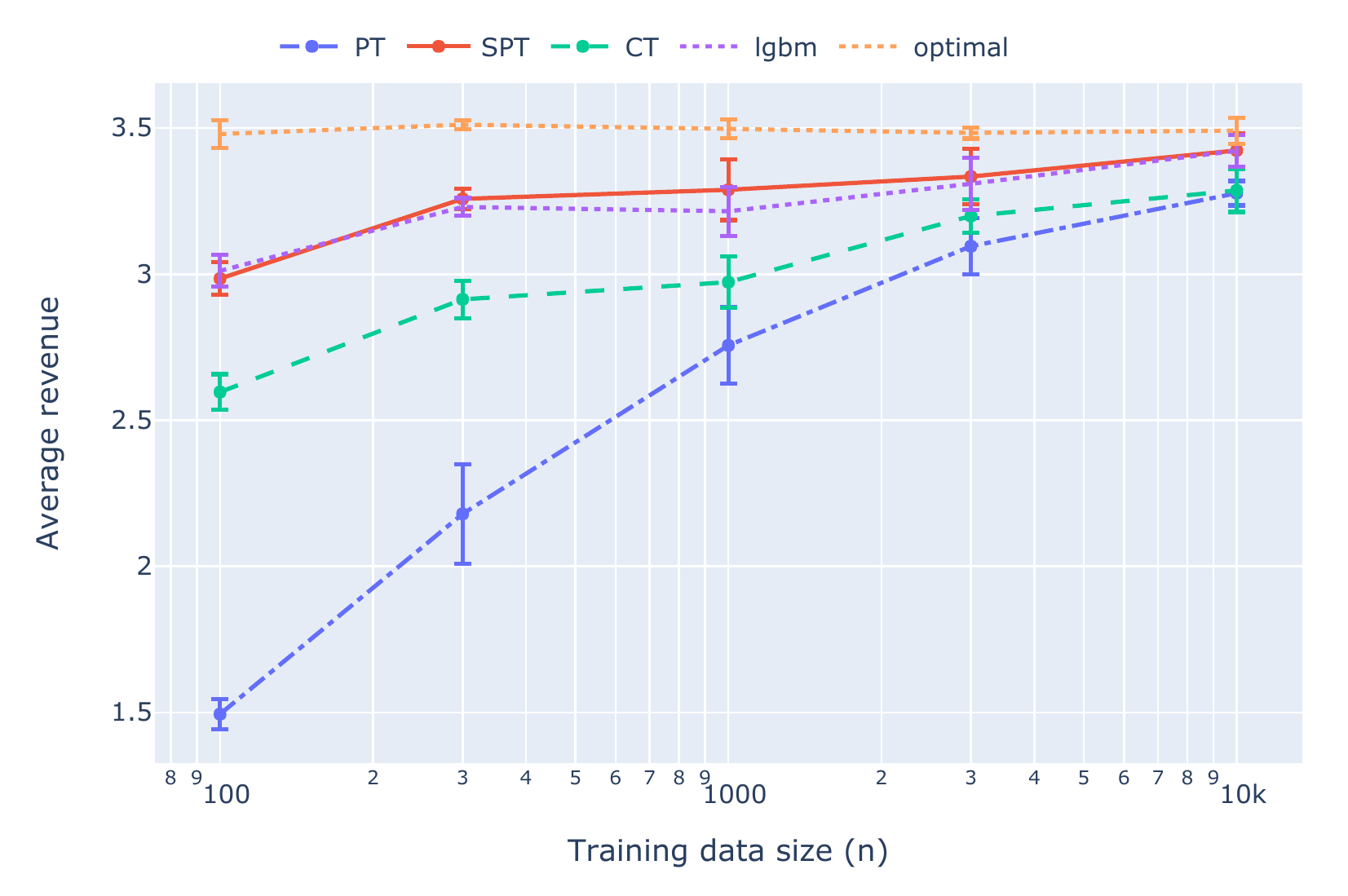}
  \caption{Revenue with varying training data size}
  \label{rev_with_datasize}
  \end{subfigure}
  \caption{Experiment results on synthetic Dataset 4}
\end{figure}
The datasets we examine come from 
the following generative model:
\begin{equation}
	Y^*=g(X)+h(X)P + \epsilon, \quad Y=
    \begin{cases}
      1, & \text{if}\ Y^*>0 \\
      0, & \text{if}\ Y^*\leq 0
    \end{cases} \label{latent_eq} \\ 
\end{equation}
\begin{itemize}
\item Dataset 1: linear probit model with no confounding: $g(X)=X_0$, $h(X)=-1$ and $X \sim N(5,I_2)$ and $P \sim N(5,1)$.
\item Dataset 2: higher dimension probit model with sparse linear interaction $g(X)=5,~ h(X)= - 1.5 (X'\beta)$,  $\{X_i\}_{i=1}^{20} , \{\beta_i\}_{i=1}^5, \epsilon_i \sim N(0,1), P_i\sim N(0,2), \{\beta_i\}_{i=6}^{20} = 0$, where the purchase probability is only dependent on the first 5 features.
\item Dataset 3: probit model with step interaction: $g(X)=5$,  $h(X)=-1.2 \mathds{1}\{X_0< -1\} -1.1 \mathds{1}\{-1 \leq X_0< 0\}  -0.9 \mathds{1}\{0 \leq X_0< 1\} -0.8 \mathds{1}\{1 \leq X_0 \}$.
\item Dataset 4: probit model with multi-dimensional step interaction: $g(X)=5$, $h(X)=-1.25 \mathds{1}\{X_0< -1\} -1.1 \mathds{1}\{-1 \leq X_0< 0\}  -0.9 \mathds{1}\{0 \leq X_0< 1\} -0.75 \mathds{1}\{1 \leq X_0 \} - 0.1\mathds{1}\{X_1< 0\}+  0.1\mathds{1}\{X_1 \geq 0\} $.
\blu
\item Dataset 5: linear probit model with observed confounding: $g(X)=X_0$, $h(X)=-1$, $X \sim N(5,I_2)$.
\item Dataset 6: probit model with non-linear interaction: $g(X)= 4|X_0+X_1|$,     $h(X)=-|X_0+X_1|$.
\end{itemize}

We set \blu $X= (X_0, X_1) \sim N(0,I_2)$, $P \sim N(X_0+5,2)$ \bla, $\epsilon \sim N(0,1)$, i.i.d. unless otherwise mentioned. \blu Dataset (1) and (5) are a common linear demand model used in the pricing literature, with and without observed confounding effects. \bla Datasets (2)-(4) and (6), have heterogeneity in treatment effects. A visualization of all datasets can be found in the appendix. All datasets except dataset (1) and (2) have confounding of observed features, where the price observed is dependent on the features of the item.
For the teacher model, we train a gradient boosted tree ensemble model using the lightGBM package \cite{ke2017lightgbm}. We use default parameter values, with 50 boosting rounds. 
The discretized price set is set to 9 prices, ranging from the $10^{th}$ to the $90^{th}$ percentile of observed prices in $10\%$ increments. 

We compare the personalization tree (PT), causal tree (CT) and student prescriptive tree (SPT) in terms of their expected revenue, along with the fully personalized policy of the teacher  model (lgbm). We also benchmark against the true optimum policy (optimal), which can be found by identifying the  price which results in the highest revenue according to the underlying probability model for each observation, and as such may vary depending on the observations in the sample.

We explore how the expected revenue changes with the depth of a tree ($k=\{1,2,3,4,5\}$). The number of training samples is held constant at $n=5000$.  
For each tree depth, we run 10 independent simulations for each dataset. Table \ref{revenue_table} summarizes the best, the worst, and the average over the simulations of varying tree depth.  We also compare their performance as the size of the dataset changes ($n=\{100,300,1000,3000,10000\}$), with the tree depth  set to $k=3$, again for 10 independent samples of each dataset. In all experiments each tree was grown to the full width for a given depth, i.e., $2^k$ leaves for a tree of depth $k$.

\begin{table*}[]
\centering
\begin{tabular}{lllllllll}
\hline
\multirow{2}{*}{minsplit} & \multicolumn{2}{c}{SPT}                               & \multicolumn{2}{c}{PT}                      & \multicolumn{2}{c}{CT}                            & \multicolumn{1}{c}{lgbm} & \multicolumn{1}{c}{OPT} \\
\cmidrule(lr){2-3} \cmidrule(lr){4-5} \cmidrule(lr){6-7} 
                          & \multicolumn{1}{c}{mean rev} & \multicolumn{1}{c}{n\_leaf} & \multicolumn{1}{c}{mean rev} & \multicolumn{1}{c}{n\_leaf} & \multicolumn{1}{c}{mean rev} & \multicolumn{1}{c}{n\_leaf$^*$ } & \multicolumn{1}{c}{mean rev}  & \multicolumn{1}{c}{mean rev} \\
                          \hline
50                        & 3.39 (0.012)            & 120.4                       & 2.29 (0.043)            & 158.2                       & 2.92 (0.020)            & 108.6                       & 3.37 (0.012)             & 3.49                    \\
150                       & 3.40 (0.014)            & 64                          & 2.55(0.040)             & 66.9                        & 3.03 (0.020)            & 49.4                        & 3.39 (0.010)             & 3.49                    \\
500                       & 3.39 (0.013)            & 24.6                        & 2.98 (0.037)            & 23.4                        & 3.23 (0.013)            & 15.6                        & 3.37(0.011)              & 3.49                    \\
1500                      & 3.37 (0.011)            & 5.4                         & 3.19 (0.043)            & 5.9                         & 3.28 (0.010)            & 4.5                         & 3.39 (0.019)             & 3.49    \\
\hline
\end{tabular}
\caption{Revenue comparison \texttt{minsplit} termination criterion for synthetic dataset 4 (n=5000)}
\label{minsplit_table}
\end{table*}

 As an illustrative example, we show the performance on dataset (4) in Figure \ref{rev_with_depth} and \ref{rev_with_datasize}, while detailed results for changing depth and dataset size for individual datasets can be found in the appendix. \bla
We observe that SPT generally improves as the tree gets deeper, whereas PT and CT appear to perform worse. Moreover, while all methods improve as the training data size increases, the student tree model significantly outperforms other methods on small sample sizes. One explanation is that when the trees are deep or the training size is small, the number of observations per action/price is also small. This increases the variance of the estimation of the outcome of that action and makes it less likely that the action with the highest average revenue is the one with the strongest signal, rather than noise. Having a teacher model can help reduce some of this noise within the data for small leaves, compared to PT and CT which only use the outcomes in that leaf.  We also observe that the SPT appears to perform relatively better than PT or CT in dataset 5 compared to dataset 1, indicating that the teacher model also helps control for observed confounding effects. We conjecture that this is due to violations of the assumption PT and CT make that each leaf is sufficiently small that they are effectively controlling for $x$ and therefore avoiding confounding effects. Observing the trees resulting from experiments, it is clear that this is not satisfied for the shallow interpretable trees we study and confounding effects lead to poor prices. Our model avoids this by using the teacher model to provide an estimate of selling probability, conditioned on $x$ (and therefore controlling for $x$). \blu Furthermore the student model (SPT) often outperforms the teacher model (lgbm), suggesting it is less prone to overfitting given inherently simpler functional form. 

We also observe that the SPT algorithm is relatively unaffected by the choice of price discretization relative to PT and CT. As the discretization becomes more granular, PT and CT are much more prone to overfitting, as the number of observations in the price bucket decreases. However, SPT seems less prone to this as the predictions from teacher model appears to be more smooth and stable, as we would expect from a fitted line vs raw observations. \bla


To show that our splitting criteria performs well with alternative termination criteria and to provide further insight into the interpretability of our approach, we show results where the termination criteria is determined by \texttt{minsplit} 
– if the number of observations at a potential split is less than a threshold, no further splits will occur on that branch. This also provides a test instance for deeper trees rather than wide trees. Table \ref{minsplit_table} summarizes the experiment results  over 10 randomly generated instances, with standard error in the parenthesis. It shows that  
our method outperforms the tree benchmarks on average, while the trees/policies are of a comparable complexity as shown by n\_leaf column. We emphasize that the ``one-vs.-all" 
causal tree is significantly less interpretable since each policy is formed by multiple trees. We also show our approach outperforms a naive implementation of a student-teacher approach, where we train a predictive teacher model that learns demand, then train a predictive student model to approximate the teacher, and lastly optimize to find a pricing policy based on the student model in the appendix.

\subsection{Dunnhumby grocery store case study}
\label{dunnhumby_case_study}

\begin{figure}[hbt!]
  \begin{subfigure}{.45\textwidth}
  \includegraphics[width=1\textwidth]{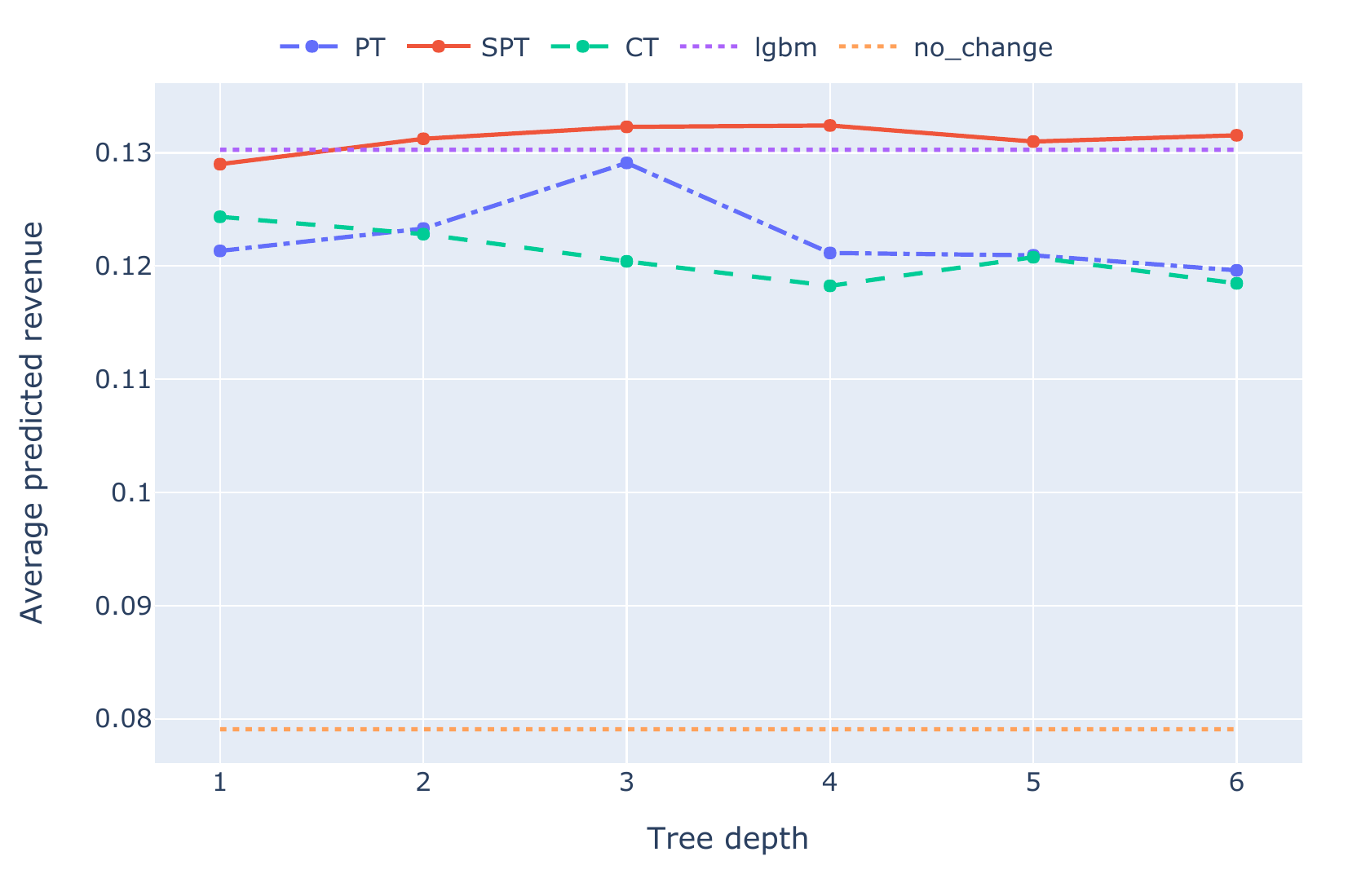}
  \caption{Predicted revenue on strawberries}
  \label{dunhumby_increasing_depth}
  \end{subfigure}
  \begin{subfigure}{.45\textwidth}
  \includegraphics[width=1\textwidth]{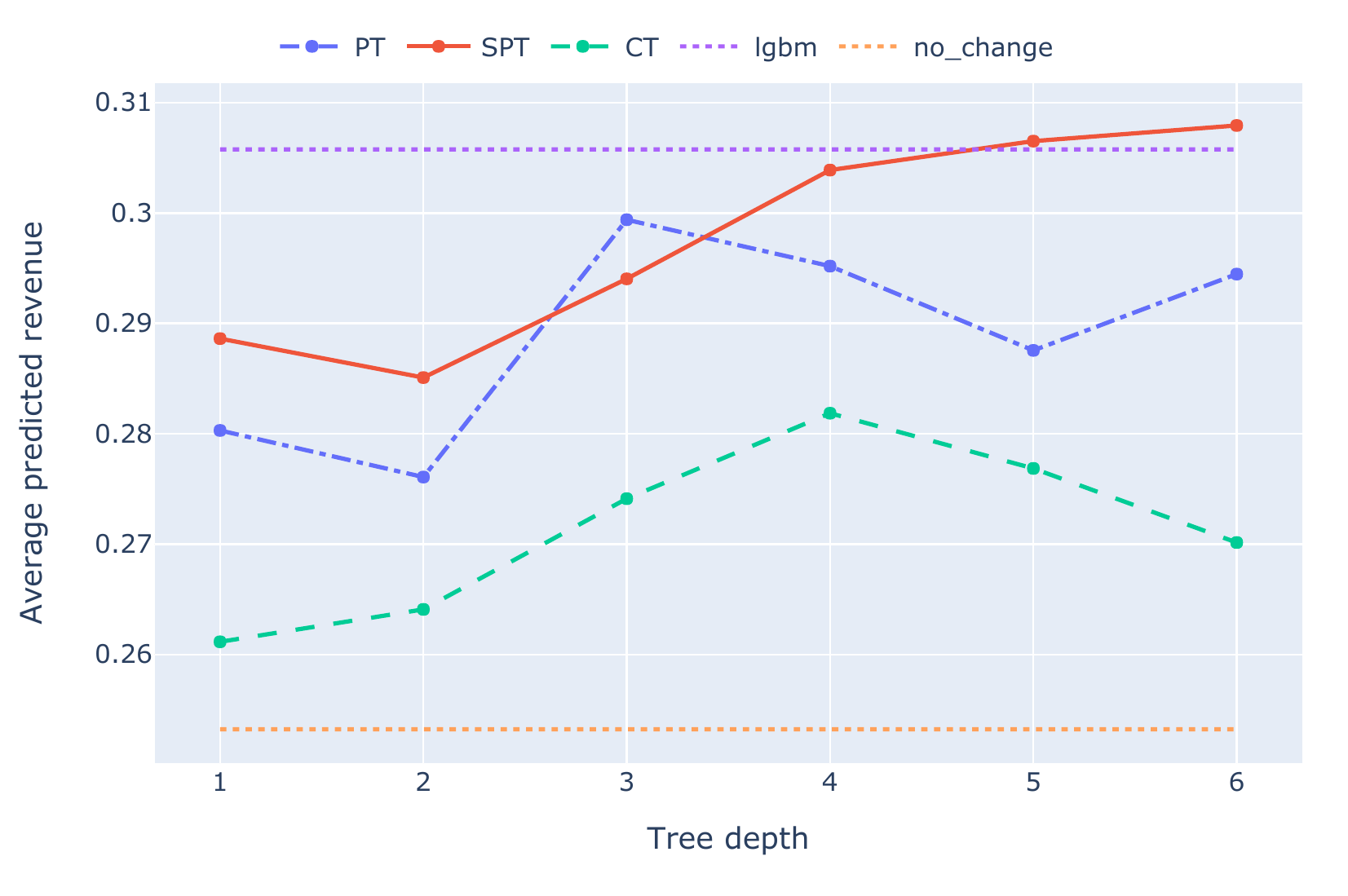}
  \caption{Predicted revenue on milk}
  \label{dunhumby_increasing_depth_milk}
  \end{subfigure}
    \begin{subfigure}{.45\textwidth}
    \includegraphics[width=1\textwidth]{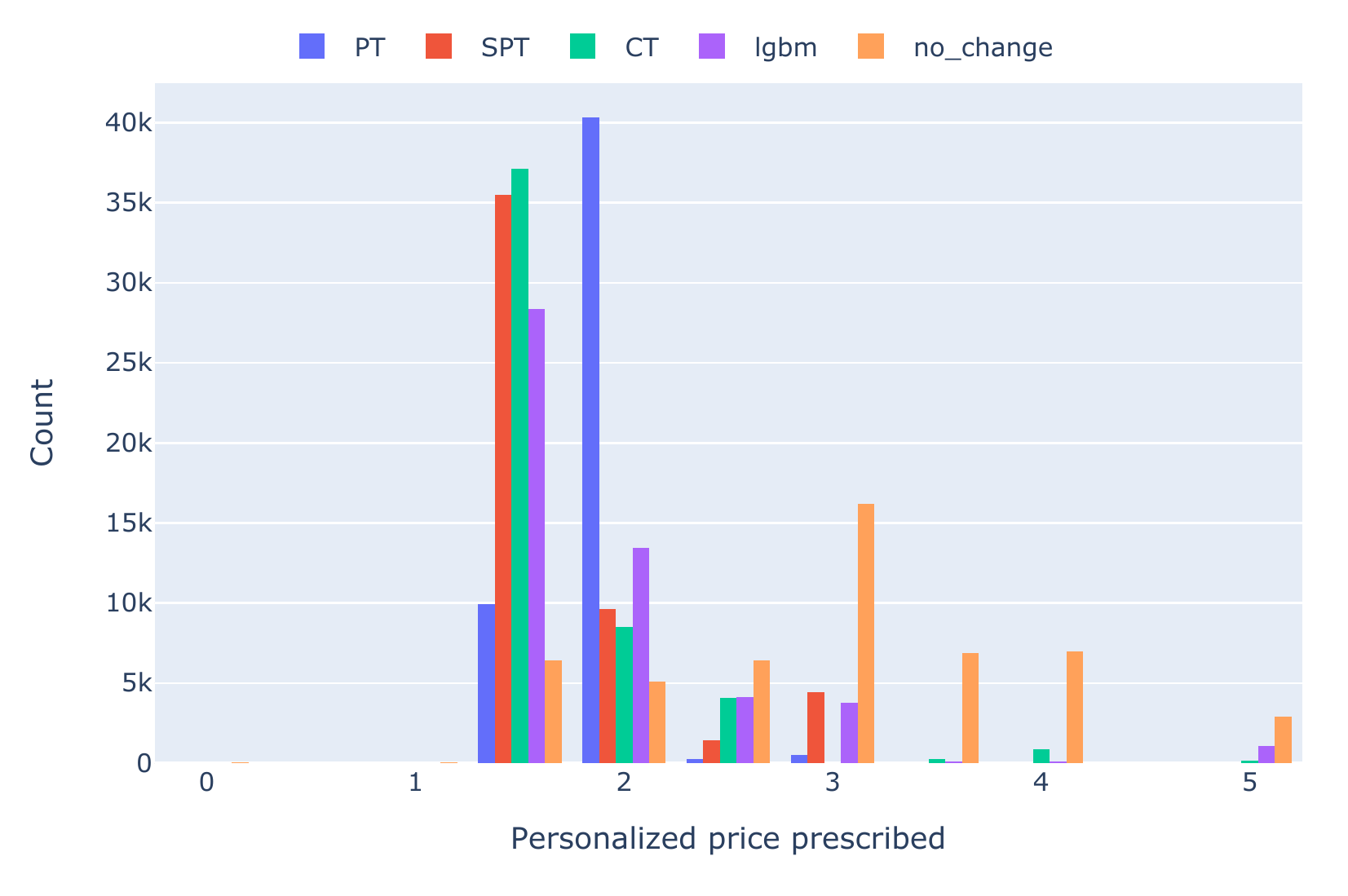}
  \caption{Prices prescribed for strawberries}
  \label{price_histogram}
  \end{subfigure}
  \caption{Experiments from Dunnhumby dataset}
\end{figure}

We benchmark the algorithms on a publicly available retail dataset collated by the analytics firm Dunnhumby.\footnote{https://www.dunnhumby.com/careers/engineering/sourcefiles} ``The complete journey'' dataset contains household level transactions over two years from a group of 2,500 households who are frequent shoppers at a grocery retailer, along with demographic information on each household. The customer features used in the \blu personalized pricing algorithm \bla are the discretized age of the shopper, discretized household income, whether they are a homeowner or renter, and the composition of the household (single male, single female, 2 adults no kids, 2 adults with kids, 1 adult with kids, or unknown). The data is processed into a format where each row corresponds to a shopping trip for a customer parameterized by their features, with a label to indicate whether the item was purchased at a listed price. \blu More details on the data pre-processing procedure can be found in the appendix. \bla We focus on strawberries (1 pound) and milk (1 gallon) for the case study due to their relatively high sales and price variation which allows the models to accurately capture the price-demand relationship. Once processed, the strawberry dataset has 102080 shopping trips, with 3373 instances where strawberries are purchased, while milk has 89936 trips with 7688 purchases. 



Due to the lack of counterfactuals in the dataset, the efficacy of the pricing algorithms are evaluated using counterfactuals estimated from an independent lightGBM model. That is, for every price prescribed, the evaluator model predicts what the selling probability will be and calculates the expected revenue. \blu We divide the dataset in half. The evaluator model is trained on a dataset which is independent from the dataset used for the teacher model and the tree-based algorithms. \bla This approach for evaluating prescriptive algorithms is used in \citealp{biggs2017optimizing}. Both the evaluator and teacher lightGBM model achieve an out-of-sample AUC of 0.79 for strawberries and 0.74 for milk with 50 boosting rounds and default classification parameters. 

Figure \ref{dunhumby_increasing_depth} and \ref{dunhumby_increasing_depth_milk} show the performance of the pricing algorithms, the teacher model (lgbm) and the current pricing policy (no\_change) in terms of predicted revenue over a range of different depth trees. SPT is able to \blu outperform the other tree approaches \bla and substantially improve upon the current pricing strategy with the predicted revenue per customer with a 67\% increase for strawberries and a 22\% improvement for milk, with a suitably chosen depth.

An example of the interpretable pricing policy for strawberries using SPT ($k=3$) is shown in Figure \ref{pricing_policy_st}. 
The model identifies that low income families are the most price sensitive and suggests a low price of \$1.49. 
 High income shoppers who are not renters are given a high price \$2.99.  \blu The distribution of the prescribes prices corresponding to different pricing policies are shown in Figure \ref{price_histogram}. \bla

\subsection{Airline pricing case study} \label{sect:airline}
\blu We worked with a large international airline and tested the prescriptive student tree method for pricing first class tickets. \bla
\blu Interpretability is paramount as airlines need to file pricing rules with Airline Tariff Publishing Company to distribute and broadcast fares over the travel service network. A potential rule may look like the following: ``if the fare is between SFO and JFK, advance purchase window is less than 7 days and the flight departs on a weekday before 11am, prescribe a price of \$400".  As regulation mandates that prices cannot differ based on customer features,  
the pricing algorithm works as an automated product differentiator by prescribing different prices to different tickets based on features such as  advance booking window, time of the  the flight, the origin and destination, distance of flight, the price of the main and basic cabin tickets, inventory level, whether the flight is one-way or round trip, etc. In total there are 65 features to train a teacher model, 
only 6 features are used to train a SPT for each route: advance purchase window, inventory levels of the main and upgrade cabins, round trip indicator, whether a flight includes a Saturday night stay, and duration of a flight. Another requirement is to limit the number of fare rules. Hence, SPTs with $k=3$ or 4 are selected for each route. 
\bla

\begin{figure}[hbt!]
  \centering
  \includegraphics[width=.45\textwidth]{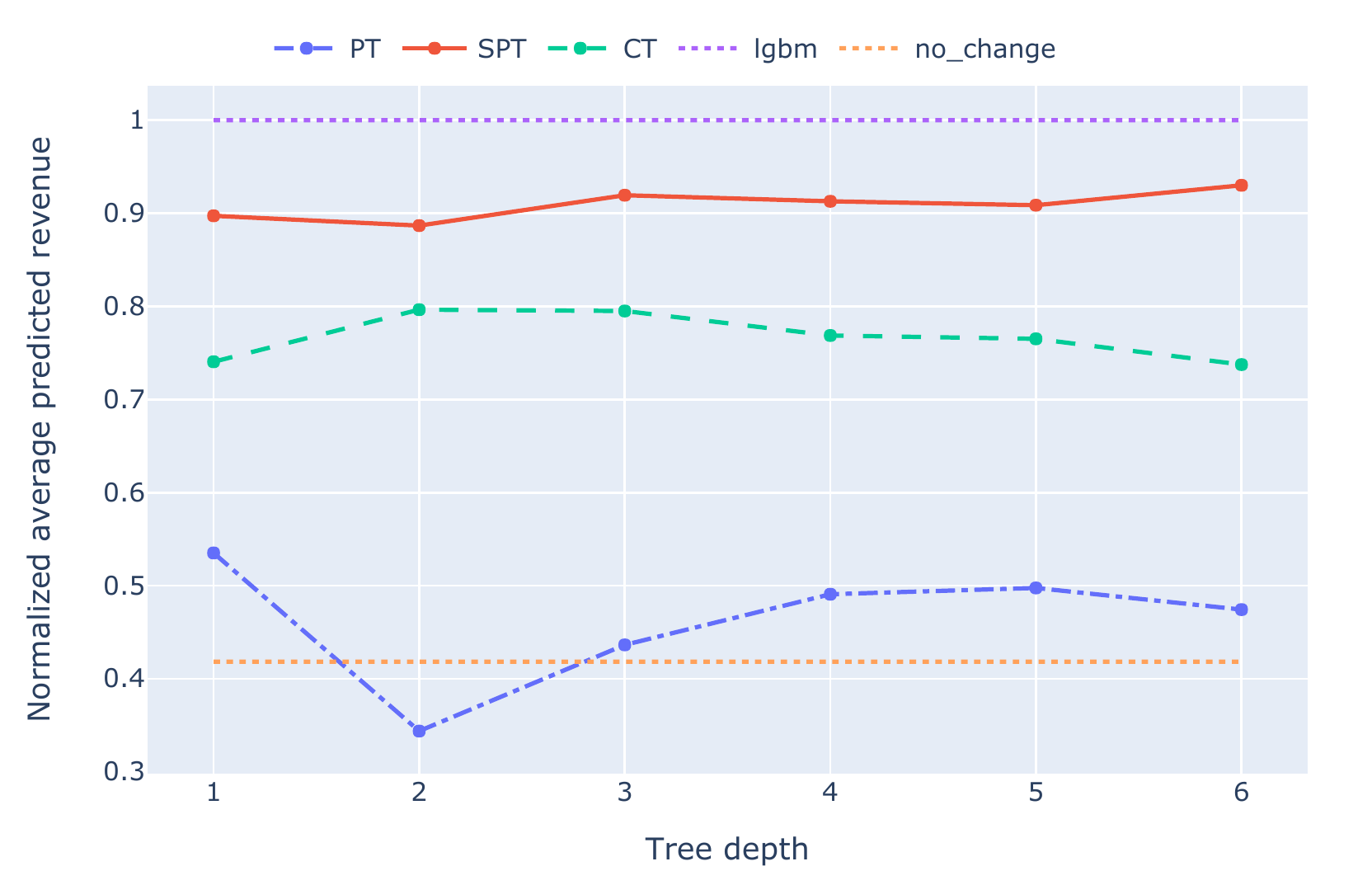}
  \caption{Predicted revenue airline pricing}
  \label{delta_increasing_depth}
\end{figure}

For confidentially reasons, some details of the study are omitted. The dataset used has 1,144,099 observations. A lightGBM model with 1000 rounds of boosting is used as the teacher model. The first class prices considered are split into \$5.00 increments over the range of prices observed in the historical data. 
Figure \ref{delta_increasing_depth} shows the predicted revenue of our proposed method to the other benchmarks, training on a 10\% (n=114,400) sample and evaluated using a lightGBM model trained on the remaining 90\% of the dataset. 
The student model is able to outperform the methods we benchmark against. Moreover, it is capable of achieving significant improvement over the current pricing with just a shallow tree.

\section{Conclusion}

\sun{Our paper focuses on a research area which has received far less attention in the literature despite its paramount practical importance, i.e., generating interpretable policies for decision making. To address this issue, we proposed a method of knowledge distillation for the  prescriptive setting by adapting the student-teacher framework. A high-capacity teacher model is incorporated to guide the student tree construction process as it estimates the counterfactual outcomes that are missing from the data, and controls for observed confounding variables. With a customized recursive partitioning algorithm, the prescriptive student tree}
is capable of simultaneously creating segments of customers with similar valuations and determining the optimal revenue-maximizing price for a segment. We quantified the performance of our algorithm in terms of the regret. Experimental results on both synthetic and real datasets showed that our approach outperforms comparable methods. Moreover, the prescriptive student tree which produced interpretable rules is capable of approximating the opaque teacher model with relatively shallow trees.

\blu While we focus on making better and interpretable pricing decisions in this paper, this framework can be easily adapted to broad prescriptive applications by replacing the splitting criterion in the student tree construction algorithm with an appropriate application-specific objective. \bla 
\sun{As techniques such as knowledge distillation in the predictive setting have enabled a proliferation of deep learning models to be deployed outside their traditional fields such as computer vision and natural language processing \cite{che2015distilling,ding2018interpreting,tang2018ranking,pan2019novel,vongkulbhisal2019unifying,liu2020general}, it is our hope that the proposed prescriptive student-teacher framework will facilitate a faster ML adoption across industries, by enabling practitioners make better business decisions and gaining more stakeholder buy-ins with interpretable explanations behind those decisions.}

\bibliography{sample-base}
\bibliographystyle{icml2021}


\end{document}


\begin{Large}
\begin{center}
    {\bf{Knowledge Distillation for Revenue Optimization:\\ Interpretable Personalized Pricing}}\\
    Supplementary Material
\end{center}
\end{Large}
~\\


\section{A Toy Example}
We use the following example to highlight the potential pitfall of a naive application of the student-teacher framework in the prescriptive setting. 

\begin{figure}[h!]
 \centering
  \includegraphics[width=4in]{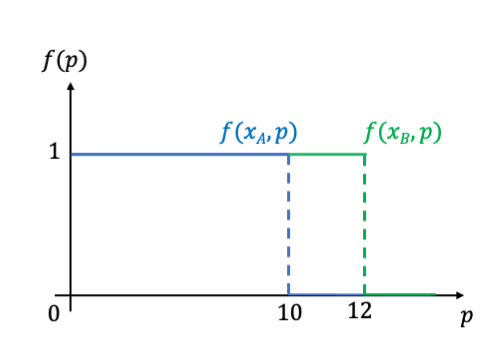}
  \caption{Response function for the toy example}
  \label{fig:toy_example}
 \end{figure}

Consider a market with an equal number of Type A and Type B customers, characterized by $x_A$ and $x_B$ respectively. Their demand or the response function $f(x,p)$ is shown in Figure~\ref{fig:toy_example}.   For Type A customers, $f(x_A, p)=1$ when $p\leq 10$, 0 otherwise; similarly for Type B customers,   $f(x_B, p)=1$ when $p\leq 12$. This is known as the perfectly inelastic demand, where demand is not influenced by a change in price (up to a maximum price). 

With a perfect teacher model, the revenue-maximizing price for Type A and B customers are simply their respective maximum valuation, i.e., $p^*_A=10$ and $p^*_B=12$, and the optimal expected revenue is  $\mathcal{R}^*=\frac{1}{2}(10)+\frac{1}{2}(12)= 11$.

Suppose only a single price is allowed. One way of naively applying the student-teacher framework here is to use a regression tree with a single leaf to approximate the optimal prices given by the teacher model by minimizing Mean Square Error, from which we obtain $p_{RT} = \frac{1}{2}(p^*_A+p^*_B)= 11$ and $\mathcal{R}_{RT} = \frac{11}{2}$. On the other hand, with the student prescriptive tree whose objective is to maximize revenue, its price and revenue are given by $p_{SPT} = 10$ and  $\mathcal{R}_{SPT} = 10$ respectively. The difference in terms of the revenue between the two methods is given by $\mathcal{R}_{RT} = \frac{11}{20}\mathcal{R}_{SPT}$. More generally, with inelastic demand given in this setup, as the maximum valuations  between the two types of customers diminish, the revenue generated by the regression tree reduces to 50\% of what is achievable with the prescriptive student tree. 

\section{Proof of Theorem 1}
\label{proof_1}
We begin by partitioning the feature space $[0,1]^d$ into a set of hypercubes of width $\frac{1}{m}$, where $m \in \mathbb{Z}^+$. Note, the furthest euclidean distance between any two points in each hypercube is $\frac{\sqrt{d}}{m}$.

An axis aligned binary tree of depth $k \in \mathbb{Z}^+$ is able to partition the feature space into any set of $2^k$ hyperrectangles. In particular find $\tilde{k} = \max k \in \mathbb{Z}^+$ such that $2^k \leq m^d$. Then $m^{-1} \leq 2^{\frac{-\tilde{k}}{d}}$.





Denote $S(x)$ as subset of observations which are in the same hypercube as $x$. For notational convenience, let $p^*=\tau^*(x)$ and  $\hat{p}=\argmax_p \sum_{i \in S(x)} p\hat{f}(x_i,p)$ be an estimation of the revenue maximizing price for hypercube which contains $x$. This is a feasible tree policy for a depth of $\tilde{k}$. The regret can be bounded as follows:

\begin{align*}
    \tau^*(x) & f(x,\tau^*(x)) -  \hat{\tau}(x)f(x,\hat{\tau}(x)) \\&=  p^*f(x,p^*) - \hat{p}f(x,\hat{p}) \\
        & = p^*f(x,p^*) - \frac{1}{|S(x)|}\sum_{i \in S(x)} \hat{p}\hat{f}(x_i,\hat{p}) + \frac{1}{|S(x)|}\sum_{i \in S(x)} \hat{p}\hat{f}(x_i,\hat{p}) -  \hat{p}f(x,\hat{p}) \\
        & \leq p^*f(x,p^*) - \frac{1}{|S(x)|}\sum_{i \in S(x)} p^*\hat{f}(x_i,p^*) + \frac{1}{|S(x)|} \sum_{i \in S(x)}\hat{p}\hat{f}(x_i,\hat{p}) -  \hat{p}f(x,\hat{p}) \\
        & \leq p^*f(x,p^*) - \frac{1}{|S(x)|}\sum_{i \in S(x)} p^* f(x_i,p^*) +  \mathcal{K}(n)\\
        & \qquad \qquad  \qquad \qquad  + \frac{1}{|S(x)|}\sum_{i \in S(x)} \hat{p}f(x_i,\hat{p}) -  \hat{p}f(x,\hat{p}) + \mathcal{K}(n) \\
        & =   \frac{1}{|S(x)|}\sum_{i \in S(x)} \left[ p^*f(x,p^*) - p^* f(x_i,p^*)\right] + \\
        & \qquad \qquad  \qquad \qquad  \frac{1}{|S(x)|}\sum_{i \in S(x)} \left[  \hat{p} f(x_i,\hat{p}) -  \hat{p}f(x,\hat{p}) \right] + 2 \mathcal{K}(n) \\
        & \leq   \frac{1}{|S(x)|}\sum_{i \in S(x)} L |x_i- x| + \frac{1}{|S(x)|} \sum_{i \in S(x)} L |x_i- x| + 2 \mathcal{K}(n) \\
        & \leq   L \frac{\sqrt{d}}{m} +  L \frac{\sqrt{d}}{m}+ 2 \mathcal{K}(n) \\
        & \leq   2 L \sqrt{d}2^{\frac{-\tilde{k}}{d}} +  2 \mathcal{K}(n) \\
        & =  2^{\frac{-\tilde{k}}{d}+1}  L \sqrt{d}+  2 \mathcal{K}(n) 
\end{align*}

The third inequality follows from optimality of $\hat{p}$ over the estimated revenue for the observations which fall in that leaf. The fourth inequality follows from the uniform bounds on the estimation error and triangle inequality. The fifth inequality follows from the Lipschitz continuity assumption and the sixth inequality follows from the maximum distance between two points in each hypercube. The final equality is due to the ability of a binary tree of depth $\tilde{k}$ to replicate the hypercubes of width $\frac{1}{m}$.

\section{Synthetic experiments}
\label{synthetic_datasets}

\begin{figure}
  \begin{subfigure}{.5\textwidth}
  \includegraphics[width=1\linewidth]{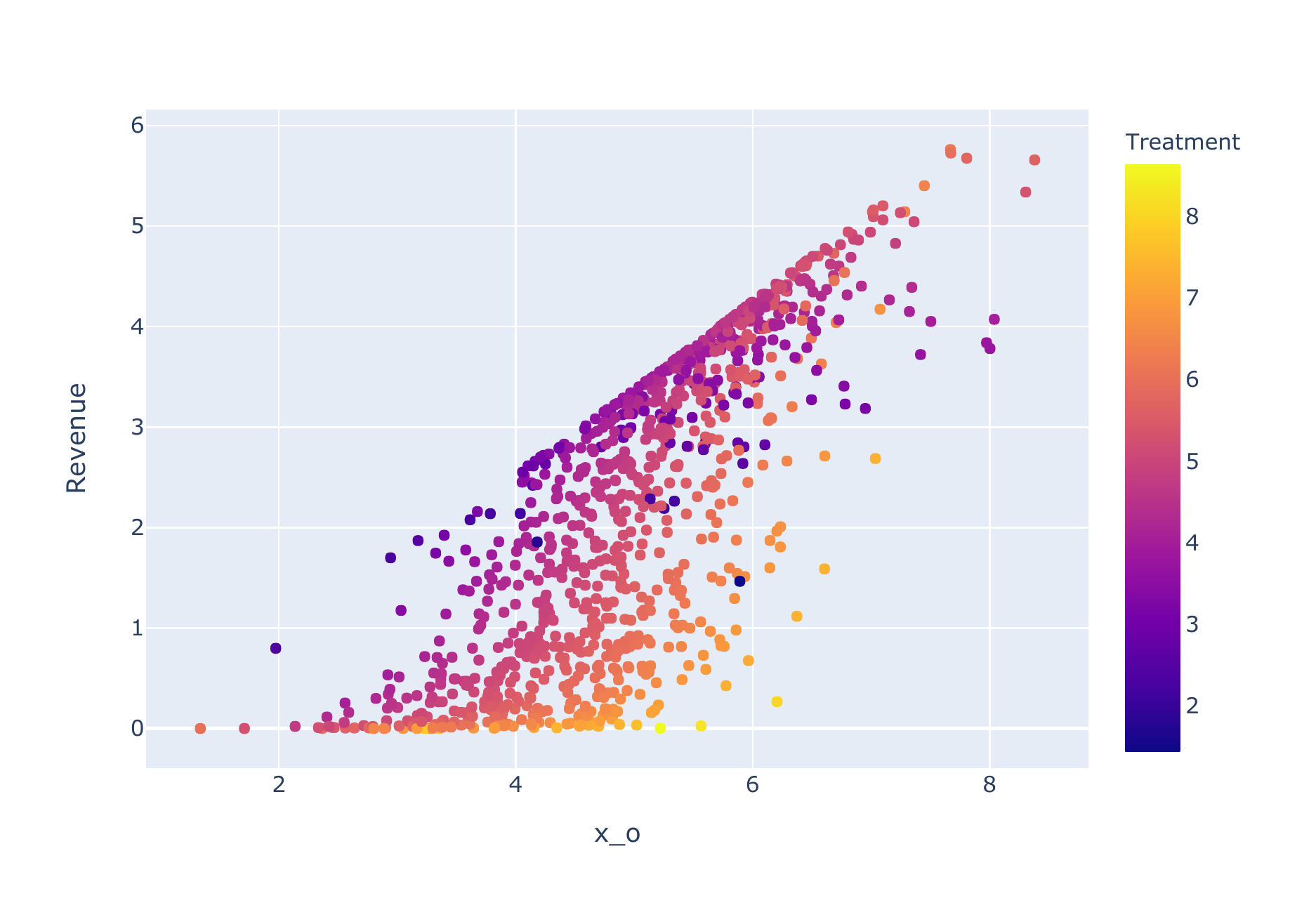}
  \caption{Dataset 1}
  \end{subfigure}
\begin{subfigure}{.5\textwidth}
  \includegraphics[width=1\linewidth]{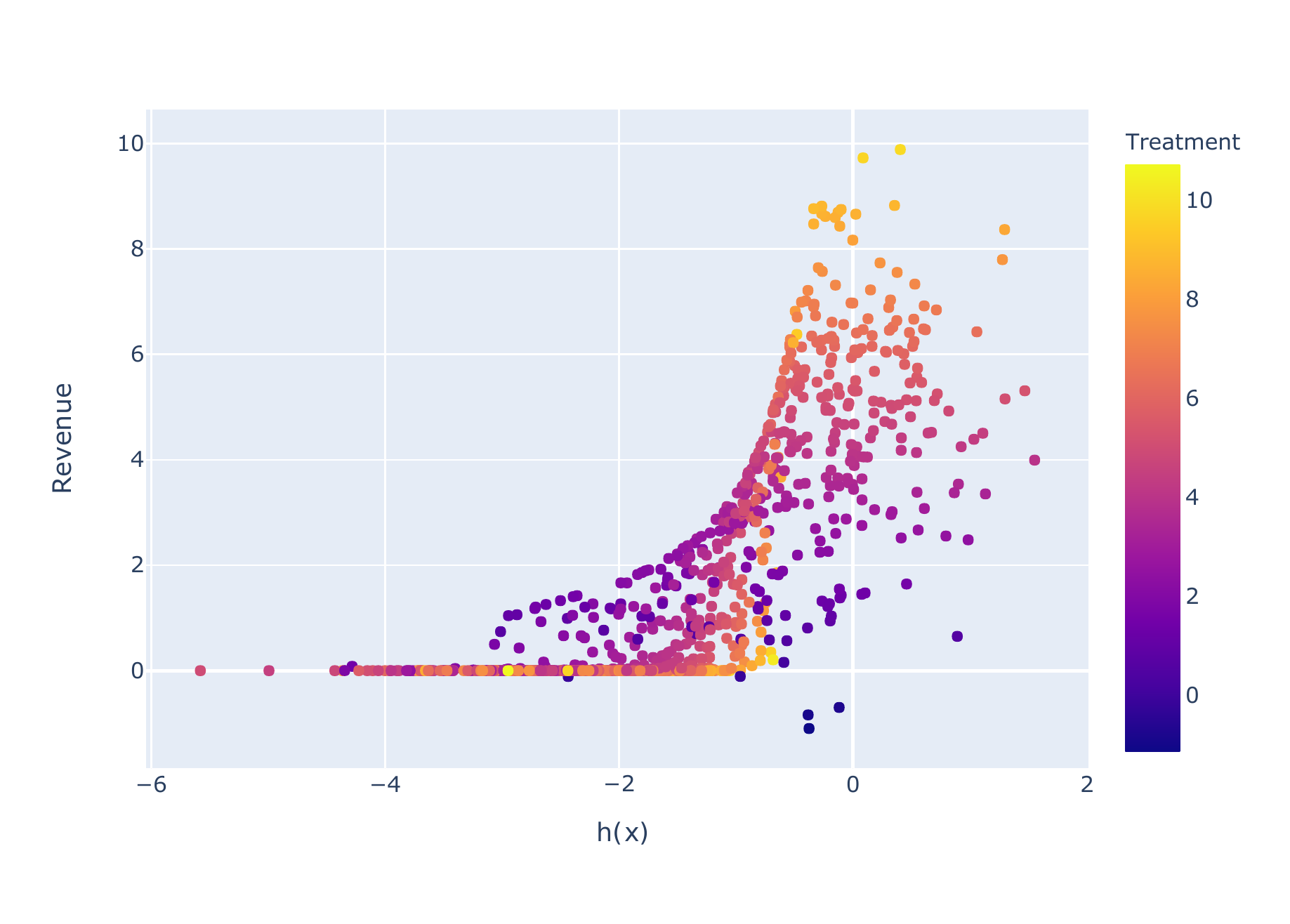}
  \caption{Dataset 2}
  \end{subfigure}
\begin{subfigure}{.5\textwidth}
  \includegraphics[width=1\linewidth]{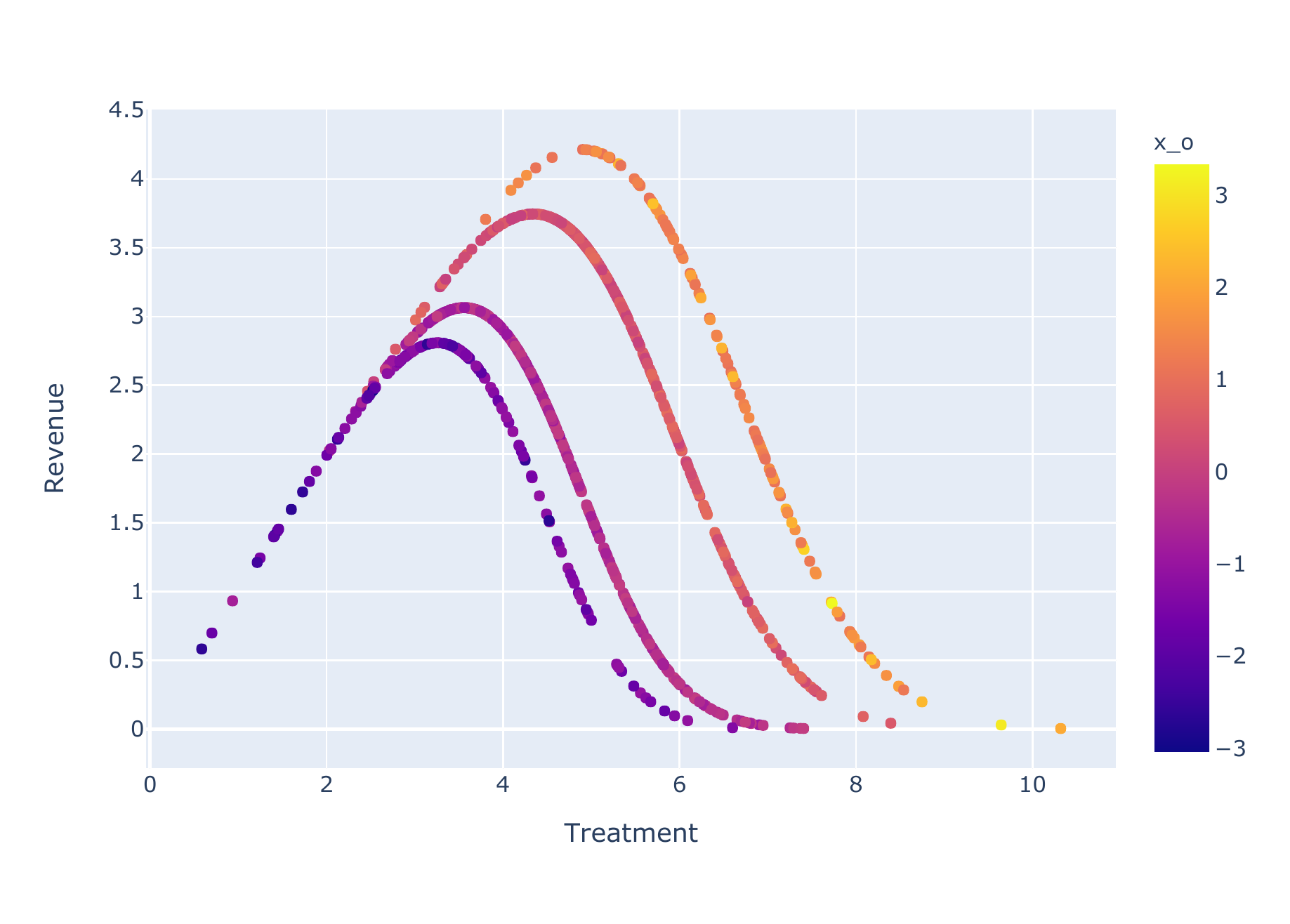}
  \caption{Dataset 3}
  \end{subfigure}
  \begin{subfigure}{.5\textwidth}
  \includegraphics[width=1\linewidth]{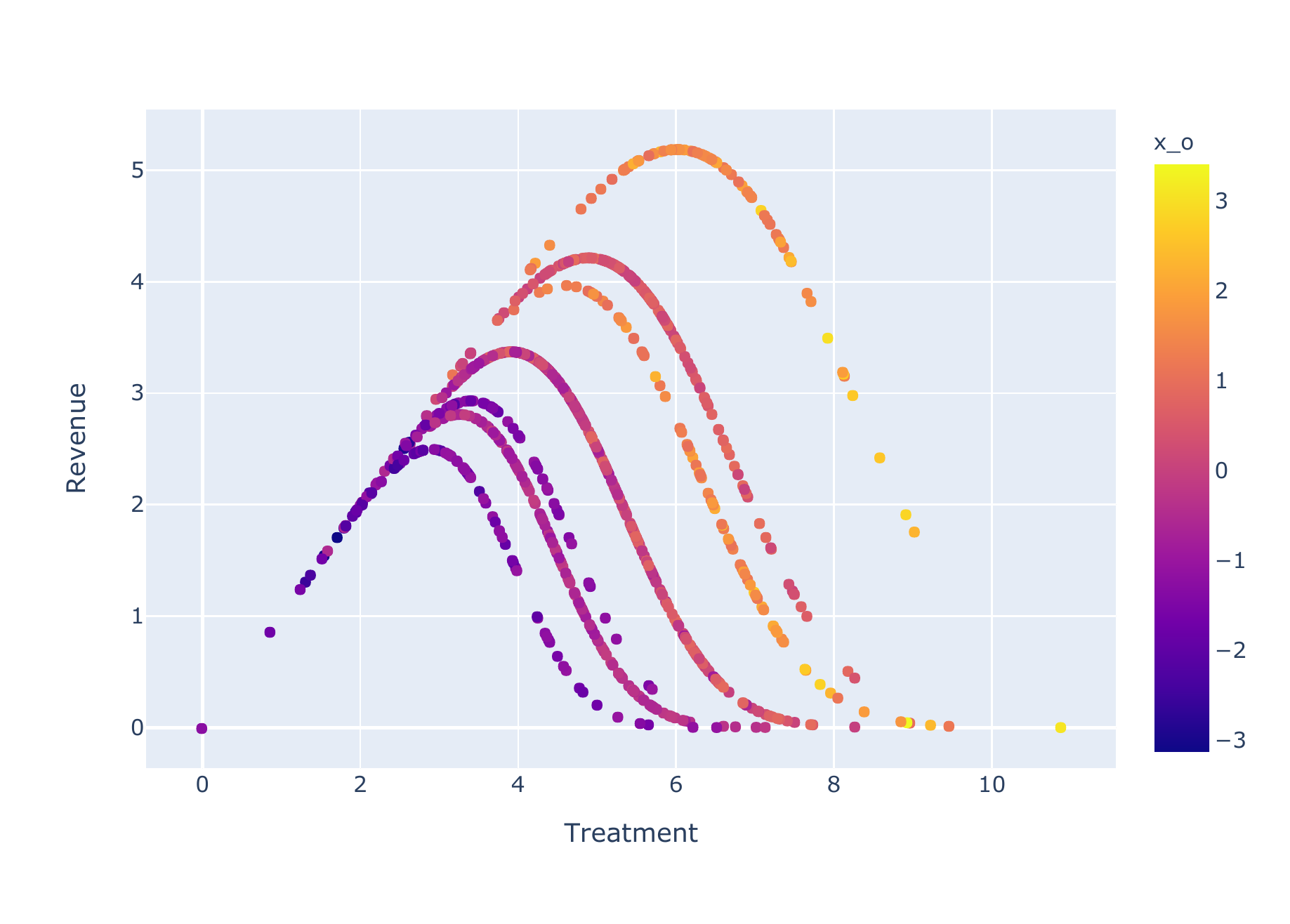}
  \caption{Dataset 4}
  \end{subfigure}
 \begin{subfigure}{.5\textwidth}
  \includegraphics[width=1\linewidth]{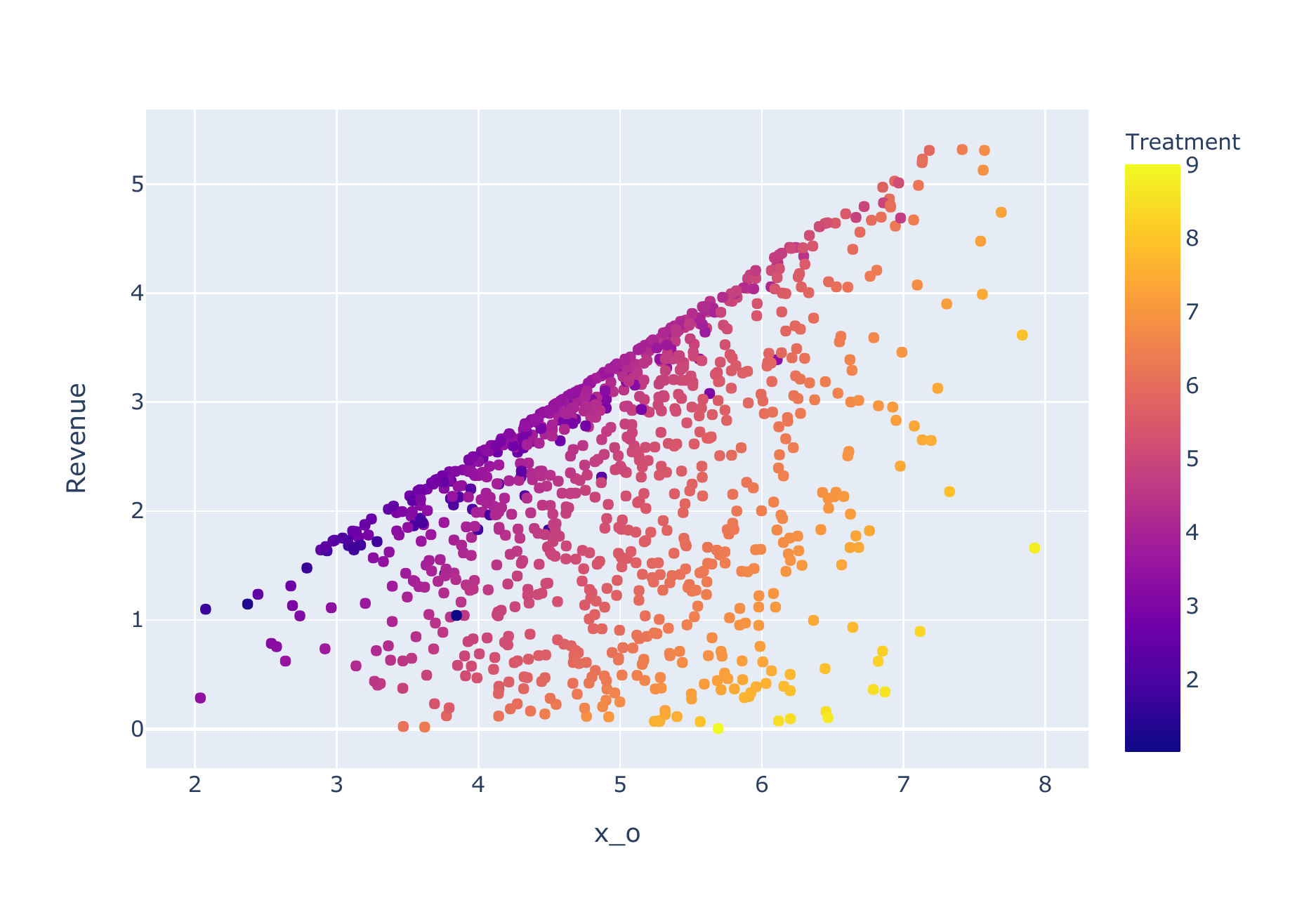}
  \caption{Dataset 5}
  \end{subfigure}
  \begin{subfigure}{.5\textwidth}
  \includegraphics[width=1\linewidth]{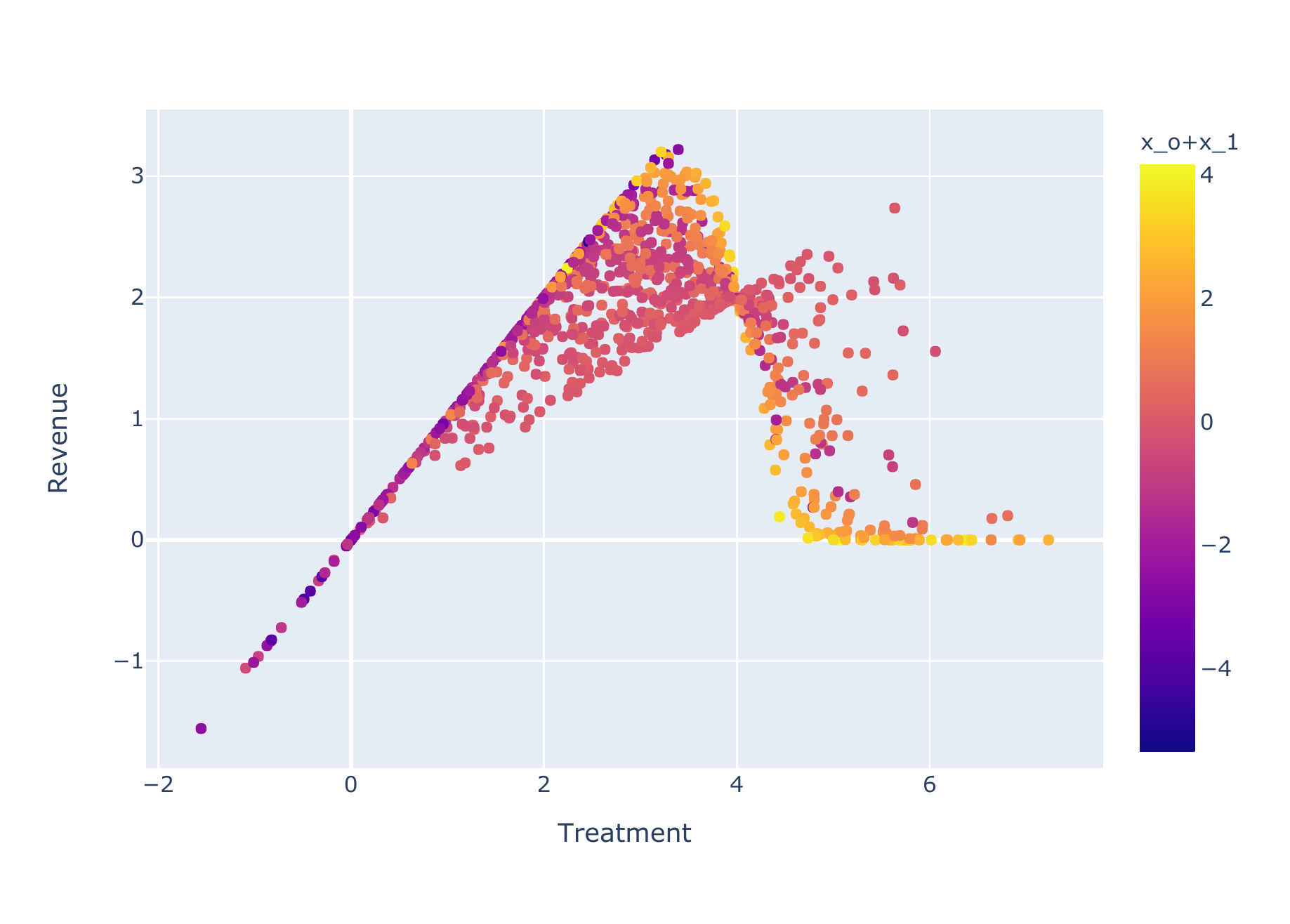}
  \caption{Dataset 6}
  \end{subfigure}
  \caption{Visualization of revenue of synthetic datasets}
  \label{fig:rev}
\end{figure}

Datasets (1)-(6) used are visualized in Figure \ref{fig:rev}.  This provides insight into the shape of expected revenue, as a function of the treatment (price), $X_i$ or $h(X)$ (the coefficient in front of price) depending on what illustrates the relationship most intuitively.

\begin{figure}
  \begin{subfigure}{.5\textwidth}
  \includegraphics[width=1.1\linewidth]{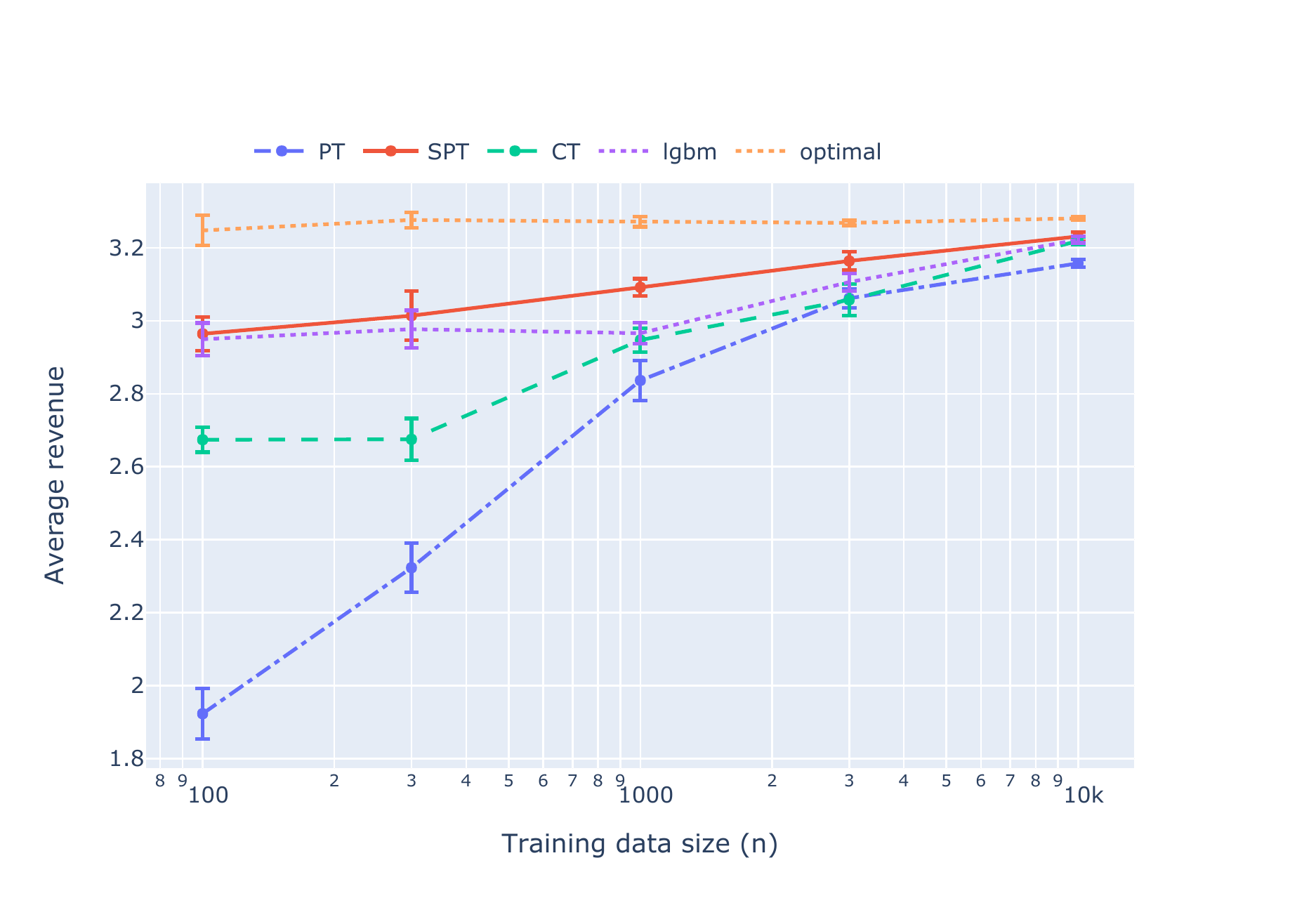}
  \caption{Dataset 1}
  \end{subfigure}
\begin{subfigure}{.5\textwidth}
  \includegraphics[width=1.1\linewidth]{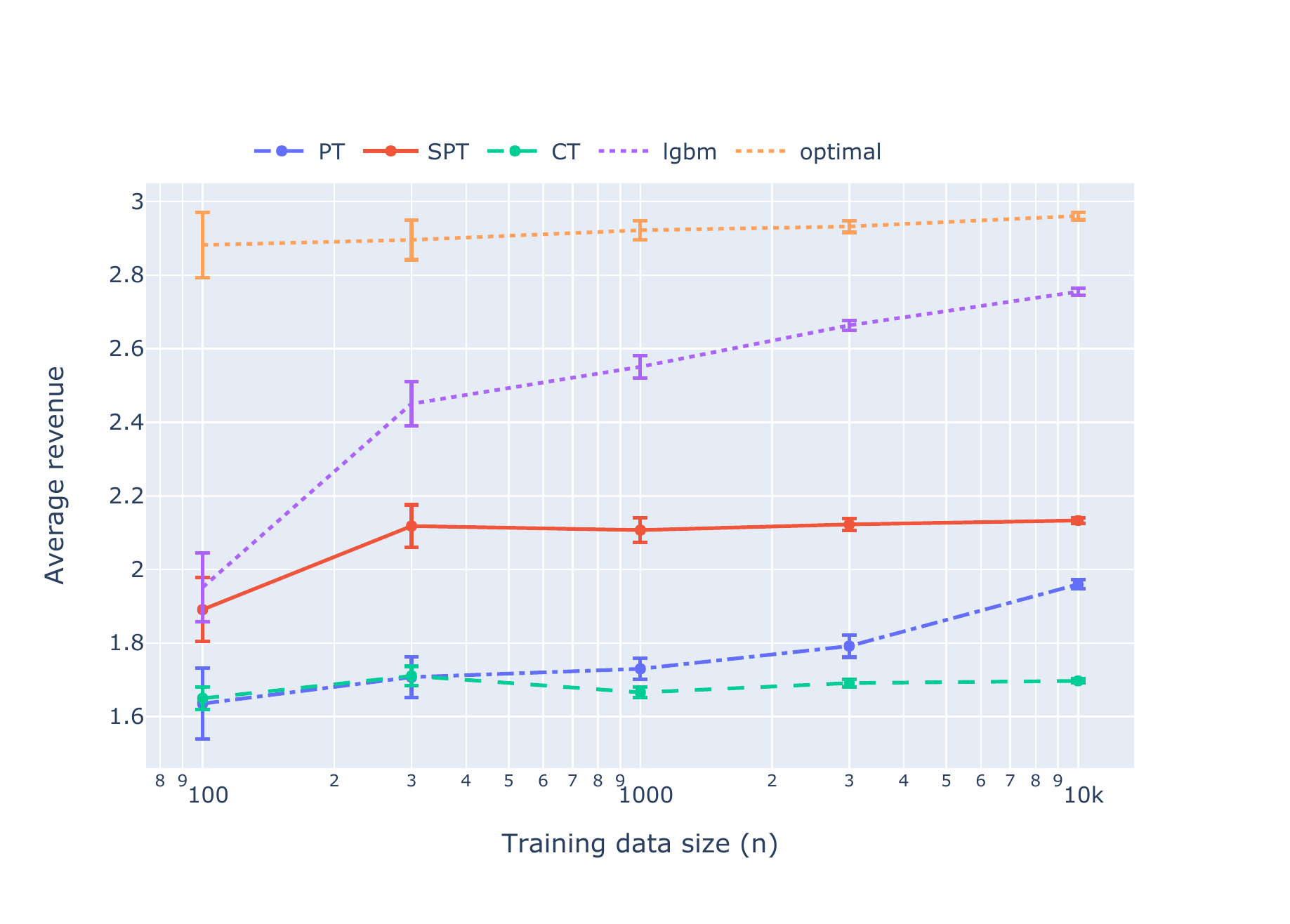}
  \caption{Dataset 2}
  \end{subfigure}
\begin{subfigure}{.5\textwidth}
  \includegraphics[width=1.1\linewidth]{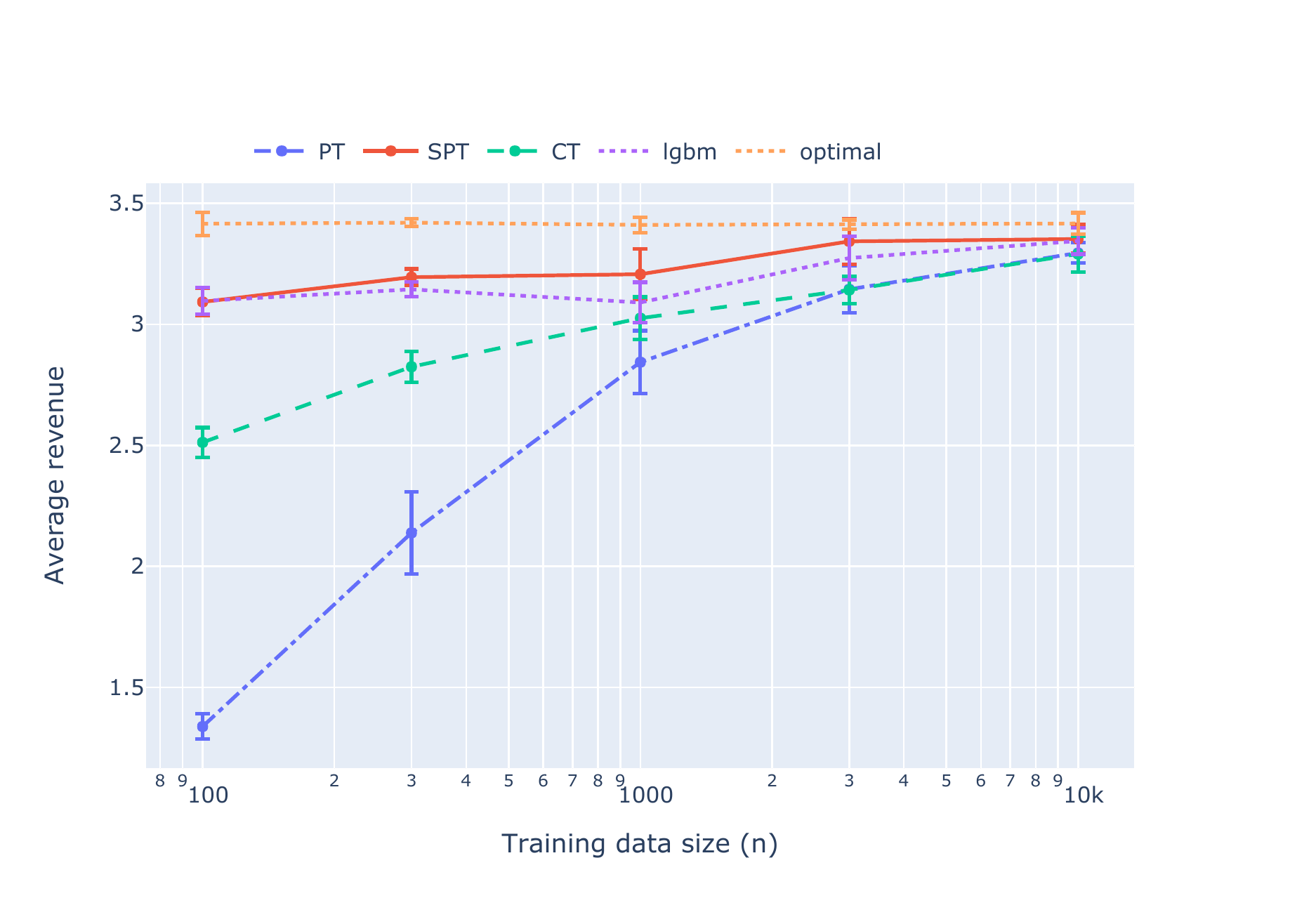}
  \caption{Dataset 3}
  \end{subfigure}
  \begin{subfigure}{.5\textwidth}
  \includegraphics[width=1.0\linewidth]{images/mean_rev_n_conf_dataset4}
  \caption{Dataset 4}
  \end{subfigure}
  \begin{subfigure}{.5\textwidth}
  \includegraphics[width=1.1\linewidth]{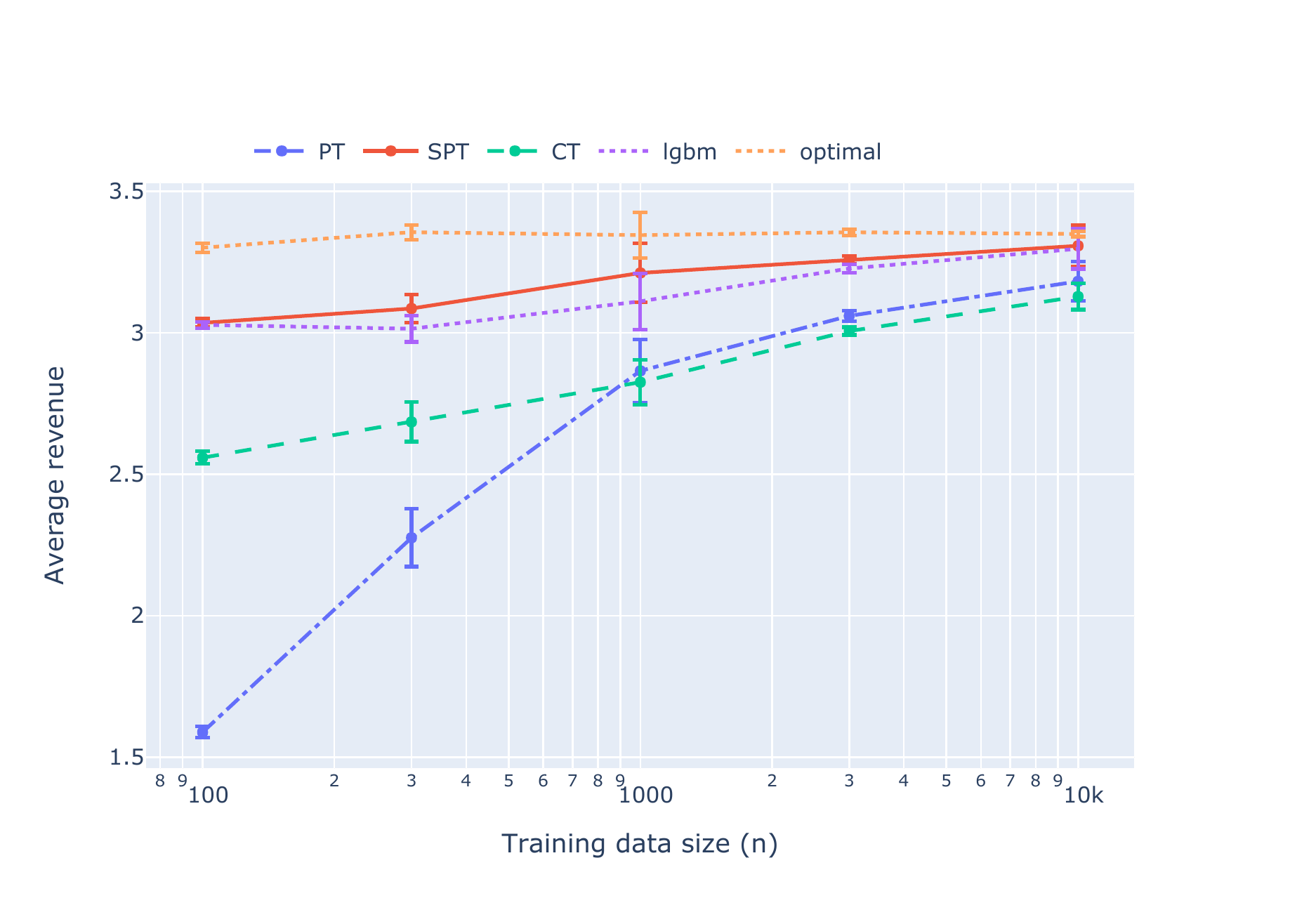}
  \caption{Dataset 5}
  \end{subfigure}
  \begin{subfigure}{.5\textwidth}
  \includegraphics[width=1.1\linewidth]{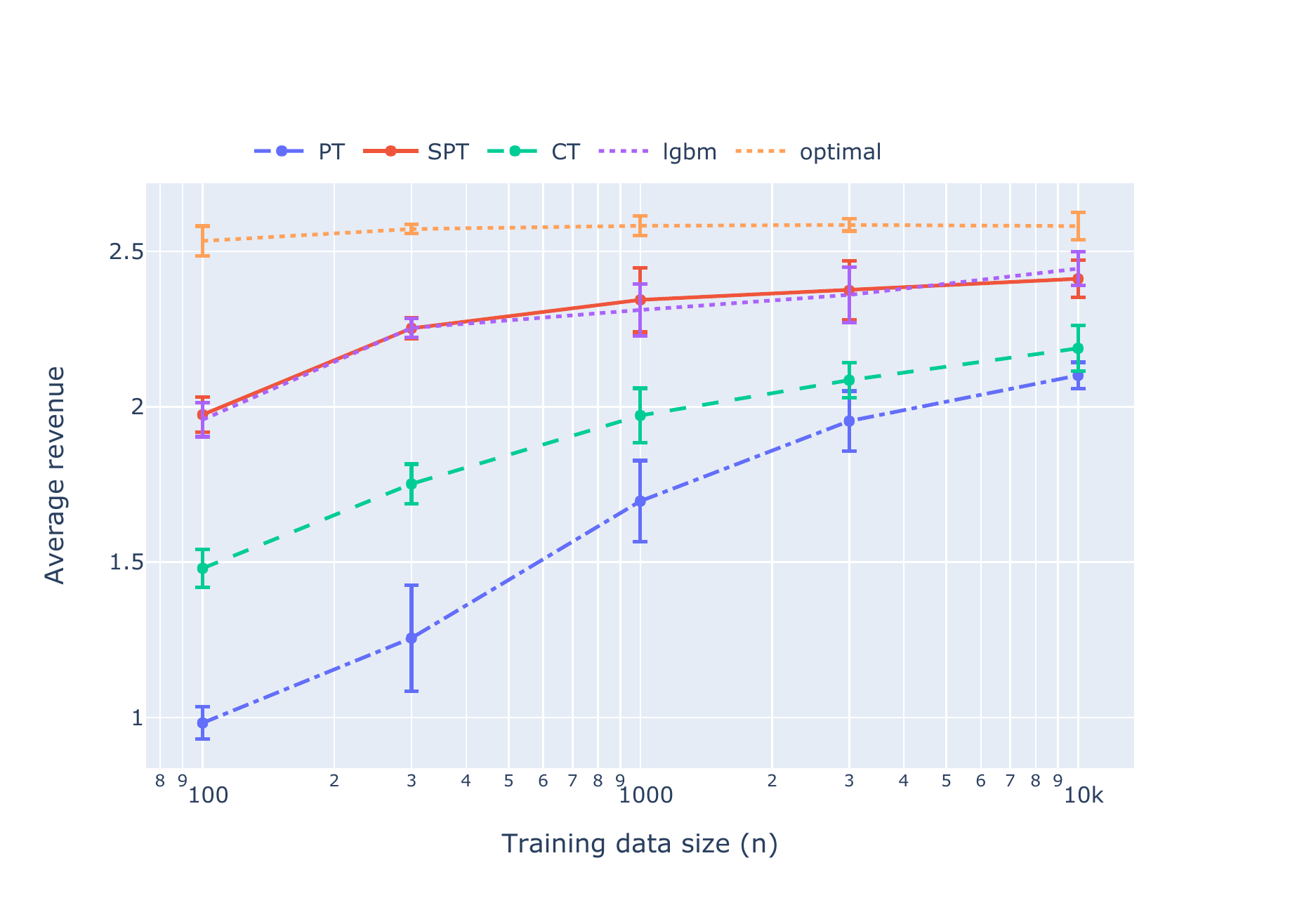}
  \caption{Dataset 6}
  \end{subfigure}
  \caption{Experiments synthetic data with with increasing data (n)}
  \label{synth_vary_n_error_bars}
\end{figure}

Figure \ref{synth_vary_n_error_bars} shows how the expected revenue changes as the size of the training sets changes for the methods we benchmark against.

\begin{figure}
  \begin{subfigure}{.5\textwidth}
  \includegraphics[width=1.1\linewidth]{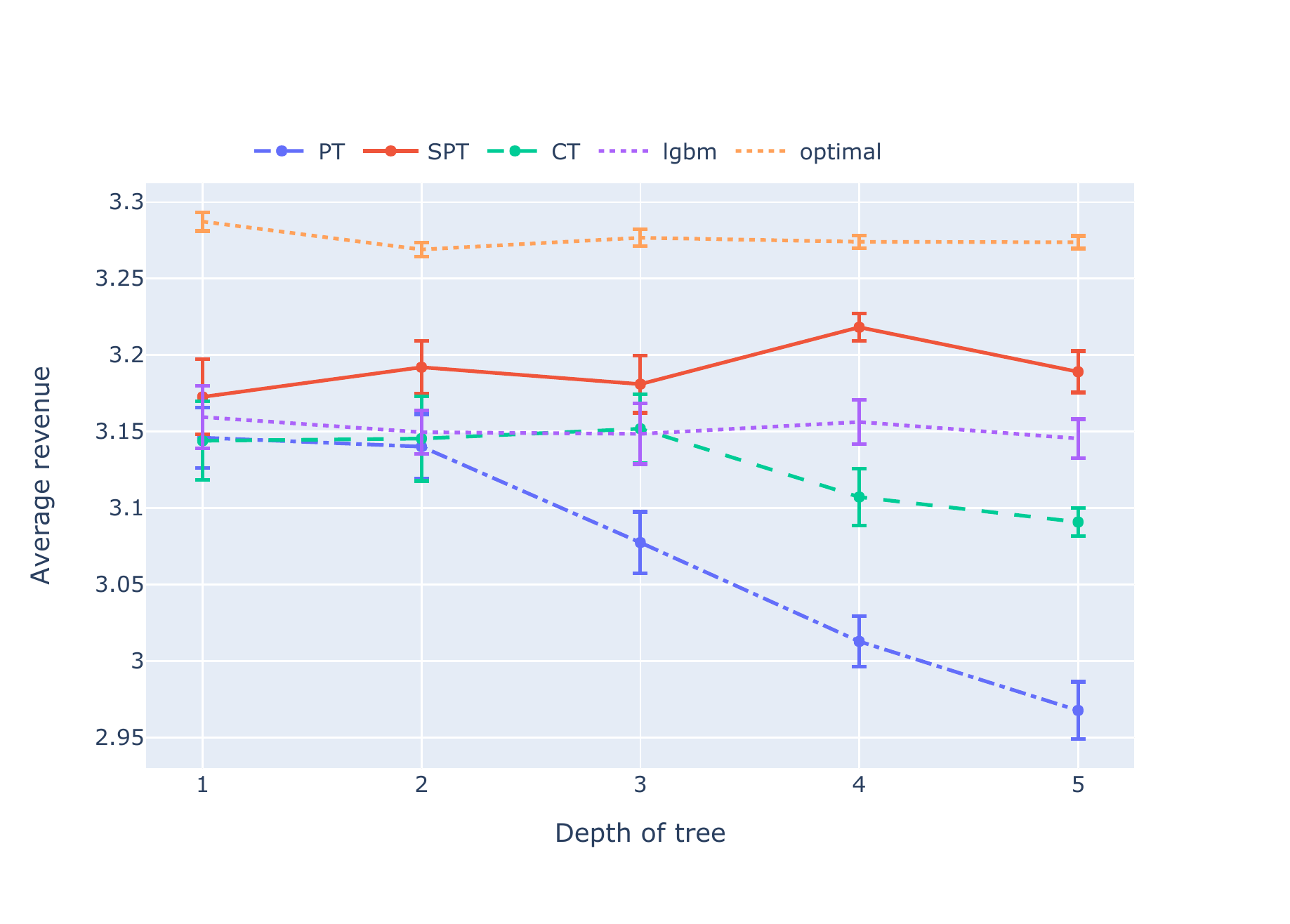}
  \caption{Dataset 1}
  \end{subfigure}
\begin{subfigure}{.5\textwidth}
  \includegraphics[width=1.1\linewidth]{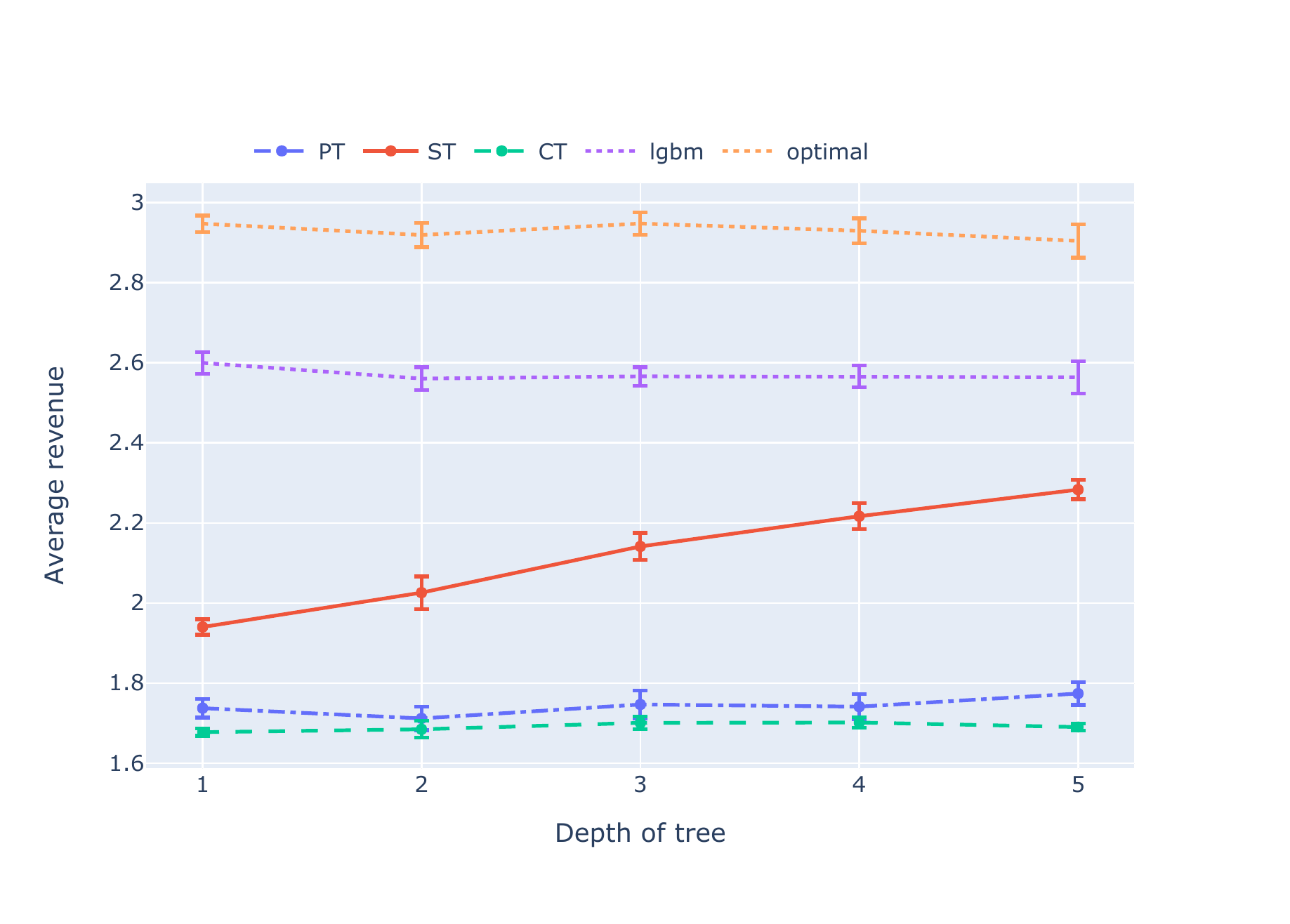}
  \caption{Dataset 2}
  \end{subfigure}
\begin{subfigure}{.5\textwidth}
  \includegraphics[width=1.1\linewidth]{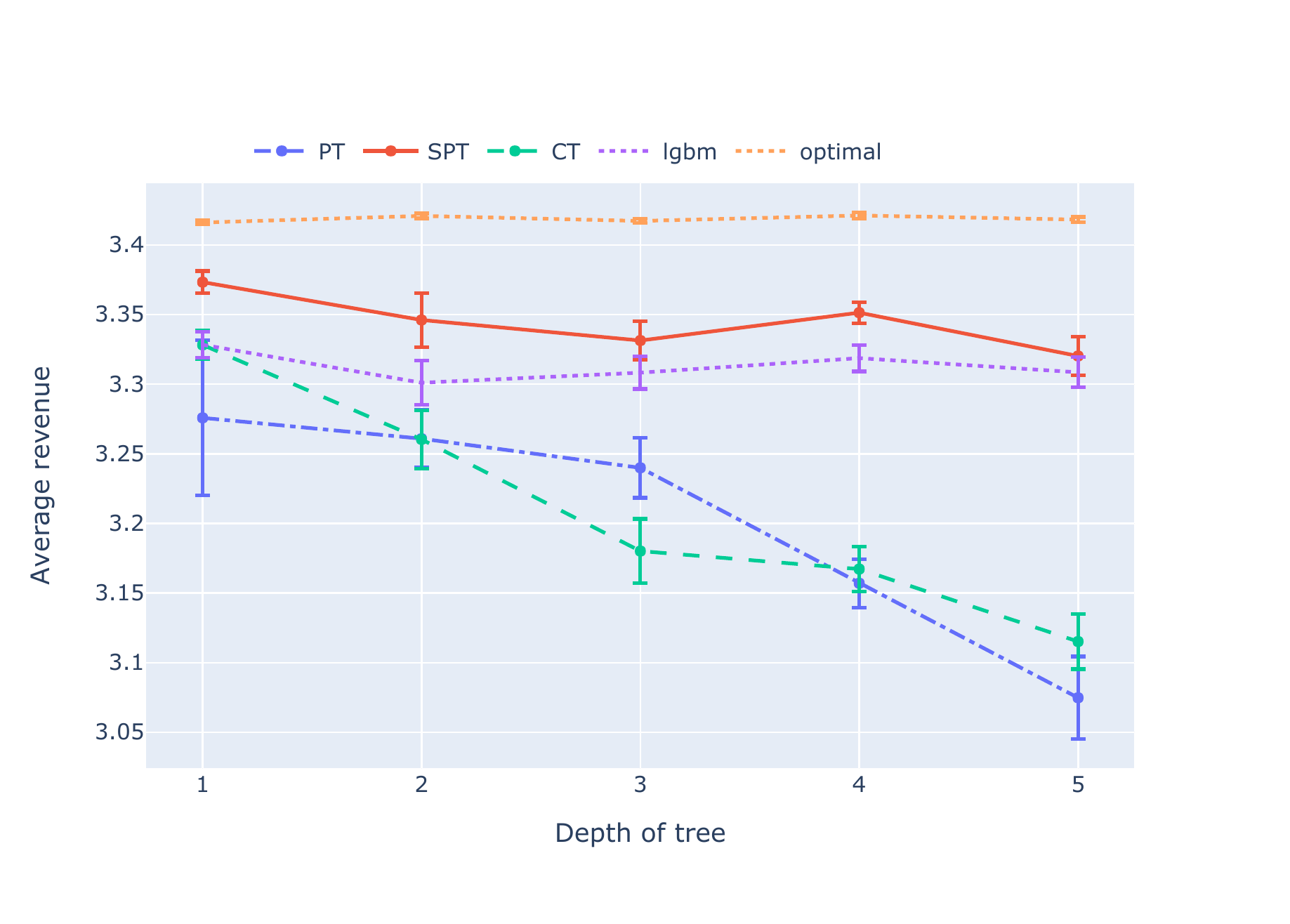}
  \caption{Dataset 3}
  \end{subfigure}
  \begin{subfigure}{.5\textwidth}
  \includegraphics[width=1.0\linewidth]{images/mean_revdepth_conf_dataset4}
  \caption{Dataset 4}
  \label{depth_datset_4}
  \end{subfigure}
 \begin{subfigure}{.5\textwidth}
  \includegraphics[width=1.1\linewidth]{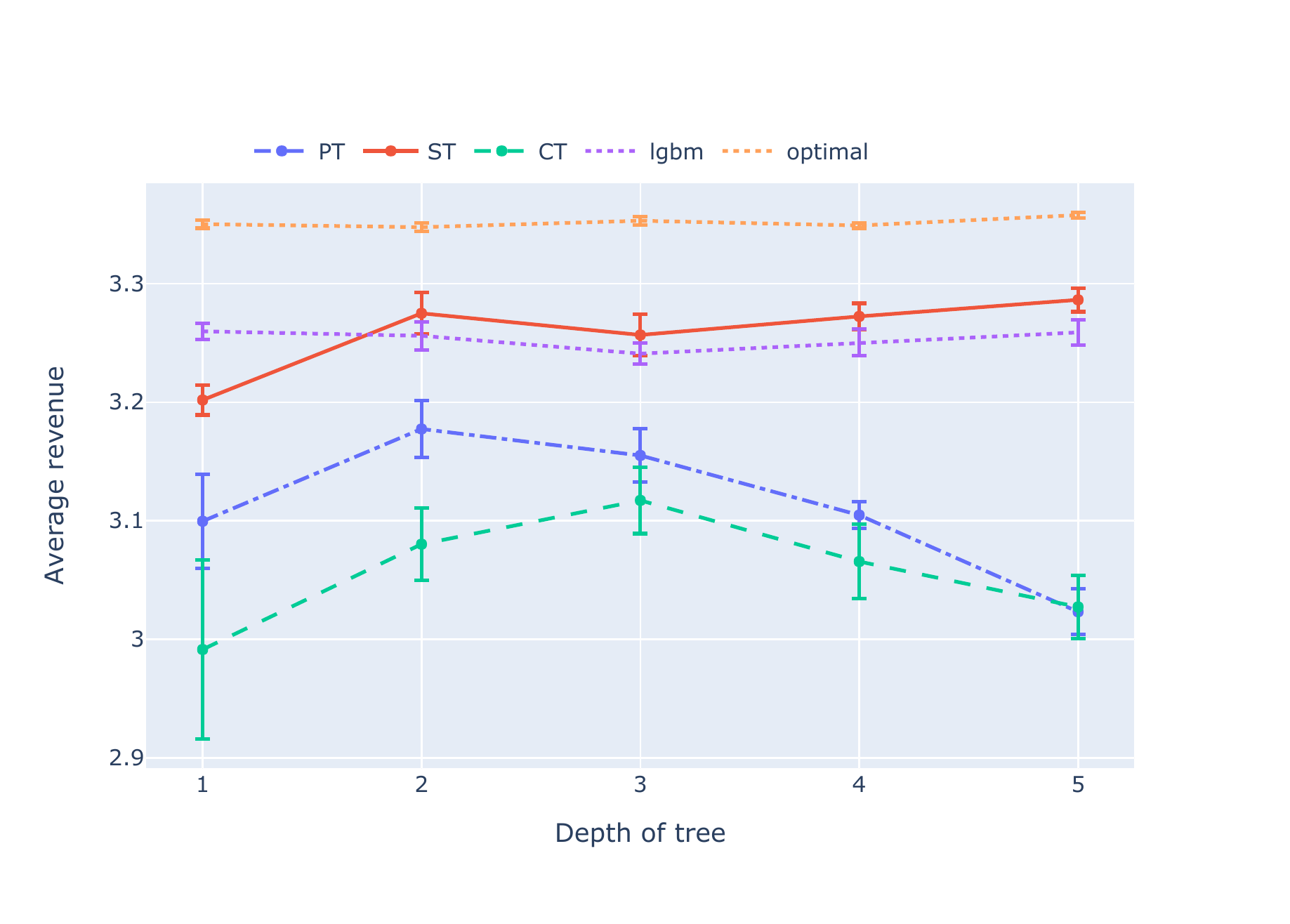}
  \caption{Dataset 5}
  \end{subfigure}
  \begin{subfigure}{.5\textwidth}
  \includegraphics[width=1.1\linewidth]{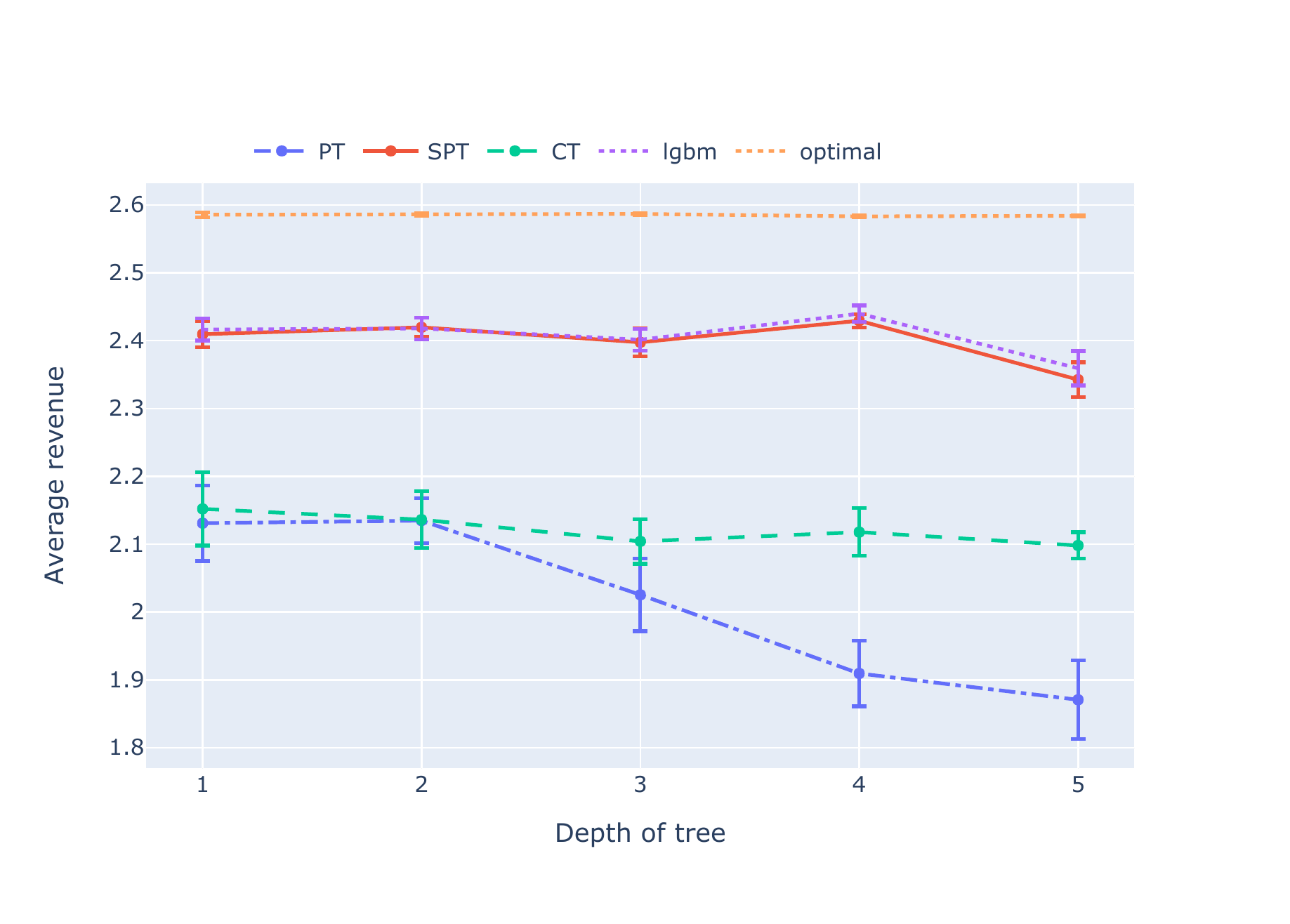}
  \caption{Dataset 6}
  \end{subfigure}
  \caption{Experiments synthetic data with increasing tree depth}
  \label{synth_vary_depth}
\end{figure}

Figure \ref{synth_vary_depth} shows how the expected revenue changes as the depth of the tree changes for the methods we benchmark against.

\textbf{Remarks on an alternative implementation of distillation} Another approach which uses the idea from knowledge distillation is as follows, train a teacher model that learns demand, then train a student model to approximate the teacher, and lastly optimize based on the student model.  We implemented the approach on Dataset (4), and revenue obtained  as tree depth increases from 1 to 10 is: 1.85, 1.84, 2.44, 2.58, 2.80, 2.98, 3.02, 3.05, 3.03, 3.04 (averaged over 10 repetitions). We observe this approach is significantly worse than the prescriptive methods we benchmark against comparing for the same depth tree (see fig 2a in paper with same experimental set-up). It appears this method needs a much greater depth (therefore impacting interpretabilty) to start achieving sensible results. However, even at depth 10, the results are inferior to even very shallow SPT trees (i.e. SPT achieves 3.4 revenue at depth 3 in Fig 2a in the main paper). Other datasets resulted in similar performance. 
We conjecture that a significant difference is that SPT finds splits which result in homogeneity of optimal price, whereas this approach finds splits which find homogeneity in predicted outcome. 

\section{Preprocessing on Dunnhumby data}
\label{dunnhumby_data_processing}
The original data is in a format where each row corresponds to an item purchased on a shopping trip. To make personalized prediction on whether an item would be purchased on a particular trip by a given shopper, the data was transformed to a format where each row corresponds a shopping trip, and a label was assigned to indicate whether the item was purchased. However, when an item is not purchased in a particular shopping trip, it's price is not recorded. Due to observed differences in the price trends over time, different imputation approaches are used for strawberries and milk. For strawberries, the price was assumed to be the mode of the price of the previous 3 sales. This is consistent with the slow moving price trends observed for strawberries where the price changed infrequently. There did not appear to be differences in pricing on a store to store level. This imputation approach has an accuracy of 83\% and is marginally more accurate than just using the previous price 80\%.
Milk on the other hand, had significant variability in the price between stores. To impute the price for milk, we used the price of the last milk sale at that store. Only stores with at least 50 sales of milk were included in the data, to avoid long time lags between purchases. This imputation approach has an accuracy of 77\% accuracy with last observation of the store, compared to 49\% using the last purchased price without conditioning on the store.

The price for strawberries follows a discrete ladder of prices (from \$1.99 to \$4.99 per unit in \$0.50 increments). The price for milk gallons follows a more irregular discrete ladder of prices, with 90\% of prices falling in the set $\{\$1.66,\$1.99,\$2.32, \$2.49, \$2.69, \$2.79, \$2.89\}$, with $\$2.32, \$2.49, \$2.69$ being the most popular prices. The personalized prices in the algorithms are restricted to the same discretization. The processed data was also restricted to shoppers who purchased strawberries or milk at least once over the 2 years studied. For strawberries, this results in a dataset with 102080 shopping trips, with 3373 instances where strawberries are purchased, while milk has 89936 trips with 7688 purchases.


